\documentclass[letterpaper, 10 pt, conference]{ieeeconf}  

\IEEEoverridecommandlockouts                              

\usepackage[utf8]{inputenc} 
\usepackage[T1]{fontenc}    
\usepackage{hyperref}       
\usepackage{url}            
\usepackage{booktabs}       
\usepackage{amsfonts}       
\usepackage{nicefrac}       
\usepackage{microtype}      

\usepackage{amsmath,amssymb,mathrsfs}
\usepackage[ruled, linesnumbered, vlined]{algorithm2e}
\usepackage{comment}
\usepackage{graphicx}
\usepackage{graphics}
\usepackage[tight]{subfigure}
\usepackage{wrapfig}
\usepackage{float}
\usepackage[table]{xcolor}
\usepackage{tabu}
\usepackage{multirow}
\usepackage{array}
\usepackage{enumerate}
\usepackage{sidecap}
\usepackage{flushend}
\usepackage{hyperref}
\usepackage{color,soul}
\usepackage{url}
\usepackage{changepage}  
\graphicspath{{./figs/}}

\let\labelindent\relax
\usepackage[shortlabels]{enumitem}
                    \setlist[enumerate, 1]{1\textsuperscript{o}}

\newtheorem{definition}{\textbf{Definition}}[section]

\usepackage{color}

\definecolor{lightblue}{rgb}{0,0.2,1}
\definecolor{black}{rgb}{0,0,0}
\hypersetup{
    colorlinks=true,%
    citebordercolor=white,%
    filebordercolor=white,%
    linkbordercolor=white,%
    urlbordercolor=white,%
    citecolor=black,%
    linkcolor=black,%
    urlcolor=black%
}

\newcounter{tecounter}
\newenvironment{spmatrix}[1]
 {\def\mysubscript{#1}\mathop\bgroup\begin{pmatrix}}
 {\end{pmatrix}\egroup_{\textstyle\mathstrut\mysubscript}}

\DeclareMathOperator*{\argmin}{arg\,min}
\DeclareMathOperator*{\argmax}{arg\,max}

\title{\LARGE \bf
Solving Markov Decision Processes with Reachability Characterization from Mean First Passage Times
}

\author{Shoubhik Debnath$^{1}$, Lantao Liu$^{2}$, Gaurav Sukhatme$^{3}$
\thanks{$^{1}$Shoubhik Debnath is with NVIDIA Corporation, Santa Clara, CA 95051, USA. E-mail:
        {\tt\small sdebnath@nvidia.com}}%
\thanks{$^{2}$Lantao Liu is with the Intelligent Systems Engineering Department at  Indiana University - Bloomington,
        Bloomington, IN 47408, USA. E-mail:
        {\tt\small lantao@iu.edu}}%
\thanks{$^{3}$Gaurav Sukhatme is with the Department of Computer Science at the University of Southern California, Los Angeles, CA 90089, USA. E-mail:
        {\tt\small gaurav@usc.edu}}%
\thanks{The paper was published in 2018 IEEE/RSJ International Conference on Intelligent Robots and Systems (IROS).}
}

\begin{document}

\maketitle

\begin{abstract} 
A new mechanism for efficiently solving the Markov decision processes (MDPs) is proposed in this paper.
We introduce the notion of {\em reachability landscape} where we use the Mean First Passage Time (MFPT) as a means to characterize the reachability of every state in the state space. 
We show that such reachability characterization very well assesses the importance of states and thus provides a natural basis for effectively prioritizing states and approximating policies.  
Built on such a novel observation, we design two new algorithms -- Mean First Passage Time based Value Iteration (MFPT-VI) and Mean First Passage Time based Policy Iteration (MFPT-PI) -- that have been modified from the state-of-the-art solution methods.    
To validate our design, we have performed numerical evaluations in robotic decision-making scenarios, by comparing the proposed new methods with corresponding classic baseline mechanisms. 
The evaluation results showed that MFPT-VI and MFPT-PI have outperformed the state-of-the-art solutions in terms of both practical runtime and number of iterations. 
Aside from the advantage of fast convergence, this new solution method is intuitively easy to understand and practically simple to implement.
\end{abstract}


\vspace{-5pt}
\section{Introduction and Related Work}

Decision-making in uncertain environments is a basic problem in the area of artificial intelligence~\cite{russell02,sigaud2013markov}, and Markov decision processes (MDPs) have become very popular for modeling non-deterministic planning problems with full observability~\cite{puterman2014markov,white1993survey}.

In this paper, we investigate the exact solution methods for solving the MDPs.We are especially interested in the MDPs that are widely applied in the artificial intelligence domain, such as the decision-theoretic planning~\cite{BoutilierDTP99}, where the agent following the optimal policy eventually enter a terminal state defined as a goal/destination.

\textbf{Contribution: } Our objective is to further improve the convergence behavior of the MDPs' solving mechanism. 
To realize that, we are introducing a new MDP solution method with a novel design, and we are showing that such design allows us to gain insights on designing efficient heuristics to speed up the convergence of the MDPs.
In greater detail, we introduce the notion of {\em reachability landscape} which is essentially a ``grid-map" measuring the {\em degree of difficulty} for each state transiting to the terminal/goal state.
By ``reachability" we mean that based on the current discrete stochastic system, how hard it is for the agent to transit from the current state to the given goal state. In other words, reachability can be viewed as the {\em reachability with regards to arriving at the terminal state}.

To compute the reachability landscape, we use the Mean First Passage Time (MFPT) which can be formulated into a simple linear system. 
We show that such reachability characterization of each state reflects the importance of this state, and thus provides a natural basis that can be used to upgrade and accelerate  both value iteration (VI) and policy iteration (PI) mechanisms, e.g., it can be utilized for prioritizing states for the standard VI process and approximating policy for the standard PI procedure.  
More specifically, we design two novel algorithms that have been modified from the state-of-the-art solution methods.
Our two algorithms include the Mean First Passage Time based Value Iteration (MFPT-VI) and the Mean First Passage Time based Policy Iteration (MFPT-PI). 
Experimental evaluations showed that both MFPT-VI and MFPT-PI are superior to the state-of-the-art methods in terms of both the practical runtime and the number of iterations needed to converge. 

\textbf{Related Work:} The basic computational mechanisms and techniques for MDPs have been well-understood and widely applied to solve many decision-theoretic planning~\cite{BoutilierDTP99,sutton1990integrated} and reinforcement learning problems~\cite{busoniu2010reinforcement, van2012reinforcement}. 
In general, the fundamental solving mechanism is to exploit the dynamic programming (DP) structure which can be computed in polynomial time.  Value iteration and policy iteration are two of the most famous and most widely used algorithms to solve the  MDPs~\cite{howarddynamic60,Bertsekas1987}.

An important related heuristic for efficiently solving MDPs is the {\em prioritized sweeping}~\cite{Moore93prioritizedsweeping}, which has been broadly employed to further speed up the value iteration process. 
This heuristic evaluates each state and obtains a score based on the state's contribution to the convergence, and then prioritizes/sorts all states based on their scores (e.g., those states with larger difference in value between two consecutive iterations will get higher scores)~\cite{parr1998generalized,wingate2005prioritization}.  
Then in the immediately next dynamic programming iteration, evaluating the value of states follow the newly prioritized order. 
The prioritized sweeping heuristic is also leveraged in our MFPT based value iteration procedure, and comparisons with baseline approaches have been conducted in our experimental section.

Important related frameworks for solving MDPs also include compact representations such as linear function representation and approximation~\cite{howarddynamic60,puterman2014markov} used in the policy iteration algorithms. 
The linear equation based techniques (detailed formulation is provided in the paper) do not exploit regions of uniformity in value functions associated with states, but rather a compact form of state features that can somewhat reflect values~\cite{boutilier2000stochastic}. 
Our method for computing the MFPT can also be formulated into a linear system. 
However, the intermediate results generated from MFPT are more direct:  they very well capture -- and also allow us to visualize -- the ``importance" of states, and can lead to a faster convergence speed which is demonstrated in the experiments.

Another relevant strategy is called {\em real-time dynamic programming (RTDP)}~\cite{Barto1995} where states are not treated uniformly. 
Specifically, in each DP iteration, only a subset of most important states might be explored, and the selection of the subset of states are usually built on and related to agent's exploration experience. 
For a single DP iteration, the RTDP usually requires less computation time in comparison to the classic DP where all states need to be swept, and thus can be extended as an online process and integrated into the real-time reinforcement learning framework~\cite{bonet2003labeled}.
Similar strategies also include the {\em state abstraction}~\cite{andre2002state,li2006towards}, where states with similar characteristics are hierarchically and/or adaptively grouped together, either in offline static or online dynamic aggregation style. 
Although we believe our proposed framework can be easily extended to the fashions of RTDP's partial states exploration and the adaptive states abstraction/computation, in this work we consider the complete and full exploration of all states, and compare with state-of-the-art methods that evaluate across the entire state space.



\definecolor{roweven}{rgb}{1,1,1}
\definecolor{rowodd}{rgb}{0.95,0.95,0.995}

\vspace{-10pt}
\section{Preliminaries}
\label{Preliminaries}

\vspace{-5pt}
\subsection{Markov Decision Processes}
\vspace{-2pt}
\begin{definition} 
A Markov Decision Process (MDP) is a tuple $M = (S,A,T,R)$, where $S = \{s_1, \cdots, s_n\}$ is a set of states and $A = \{a_1, \cdots, a_n\}$ is a set of actions. The state transition function $T : S \times A \times S \rightarrow [0,1]$ is a probability function such that $T_{a}(s_1,s_2)$ is the probability that action $a$ in state $s_1$ will lead to state $s_2$, 
and $R : S \times A \rightarrow \mathbb{R}$ is a reward function where  $R_a(s, s')$ returns the immediate reward received on taking action $a$ in state $s$ that will lead to state $s'$.
\end{definition}

A Markov system is defined as {\em absorbing} if from every non-terminal state it is possible
to eventually enter a terminal state such as a goal/destination state~\cite{boutilier2000stochastic}. We restrict our attention to absorbing Markov systems so that the agent can arrive and stop at a goal.

 A {\em policy} is of the form $\pi = \{s_1 \rightarrow a_{1}, s_2 \rightarrow a_{2},\cdots, s_{n} \rightarrow a_{n} \}$. We denote $\pi[s]$ as the action associated to state $s$. 
If the policy of a MDP is fixed, then the MDP behaves as a Markov chain~\cite{kemeny1959finite}.

\vspace{-5pt}
\subsection{MDP Solution Methods}
\vspace{-3pt}
Here we discuss prevalent solution methods including the value iteration and policy iteration, as well as state-of-the-art heuristics/variants. 

\subsubsection{Value Iteration}
The value iteration (VI) is an iterative procedure that calculates the value (or utility in some literature) of each state $s$ based on the values of the neighbouring states that $s$ can directly transit to. 
The value $V(s)$ of state $s$ at each iteration can be calculated by the Bellman equation shown below
\begin{equation} \label{eq:mdp1}
V(s) = \max_{a\in A} \sum_{s'\in S} T_a( s, s') \Big(R_a(s, s')  + \gamma V(s') \Big),
\end{equation}
where $\gamma$ is a reward discounting parameter. The stopping criteria for the algorithm is when the values calculated on two consecutive iterations are close enough, i.e., 
$\max_{s \in S} |V(s) - V'(s)| \leq \epsilon$, 
where $\epsilon$ is an optimization tolerance/threshold value, which determines the level of convergence accuracy.


\textbf{Prioritized Sweeping: } The standard Bellman recursion evaluates values of all states in a sweeping style, following the index of the states stored in the memory. 
To speed up the convergence, a heuristic called {\em prioritized sweeping}  has been proposed and widely used as a benchmark framework for non-domain-specific applications. 
The algorithm labels a state as more ``interesting" or more ``important" during a particular iteration, if the change in the state value is higher when compared to its previous iteration. 
The essential idea is that, the larger the value changes, the higher impact that updating that state will change its dependent states, thereby taking a larger step towards convergence. 



\begin{figure*}[t]
  \centering
  \subfigure[]
        {\label{fig:SI_Env}\includegraphics[height=1.0in]{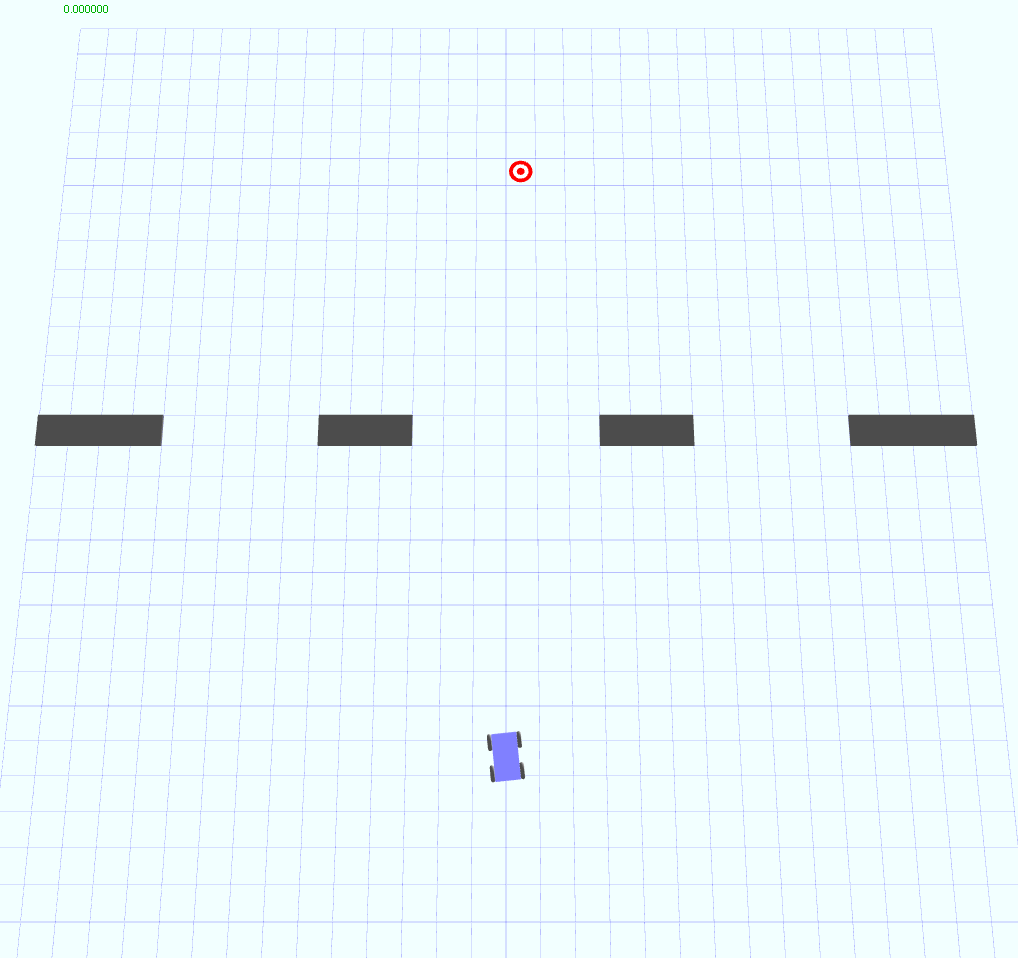}}
        \quad \quad 
  \subfigure[]
        {\label{fig:heatmap0}\includegraphics[height=1.0in]{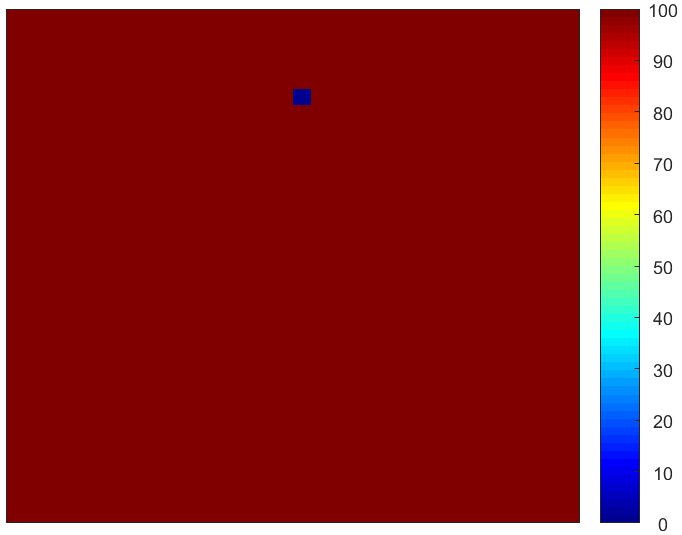}}
        \quad \quad 
  \subfigure[]    
        {\label{fig:heatmap1}\includegraphics[height=1.0in]{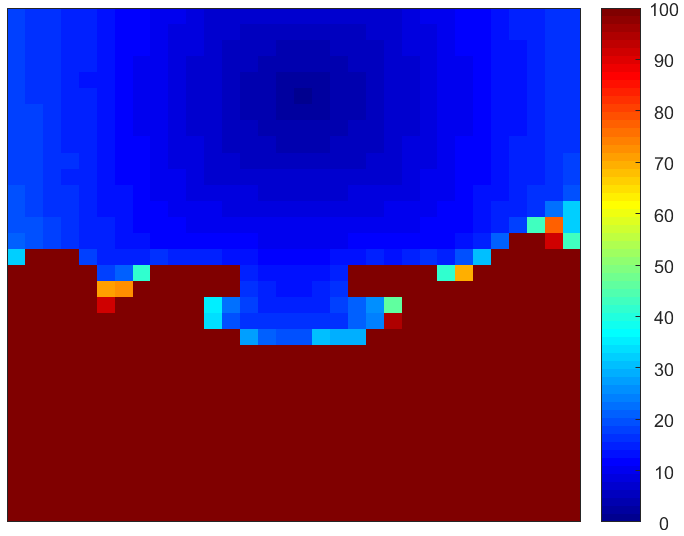}} \\
  \subfigure[]
        {\label{fig:heatmap2}\includegraphics[height=1.0in]{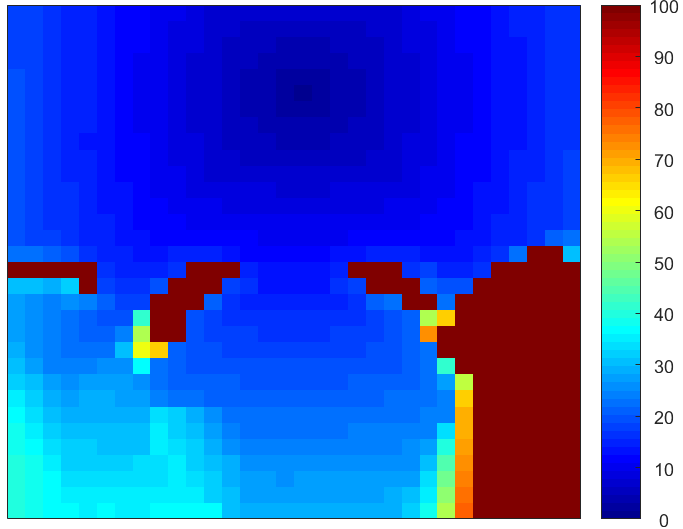}}  
        \quad \quad 
  \subfigure[]
        {\label{fig:heatmap3}\includegraphics[height=1.0in]{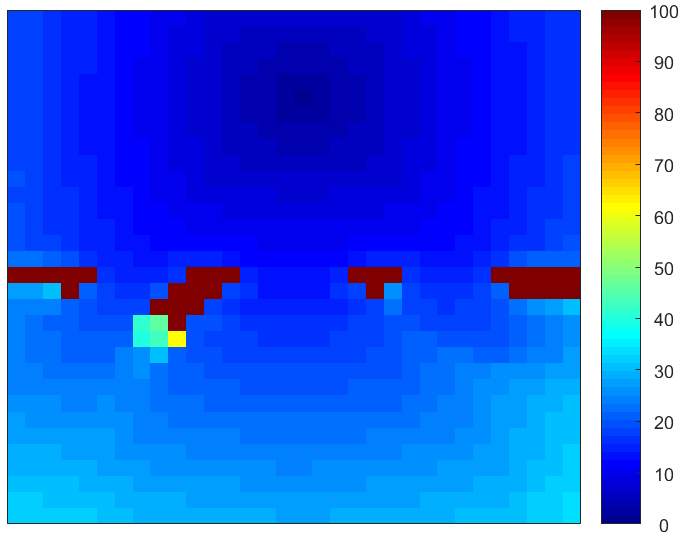}}
        \quad \quad 
  \subfigure[]
        {\label{fig:MFPT-VI_epsilon_8}\includegraphics[height=1.0in]{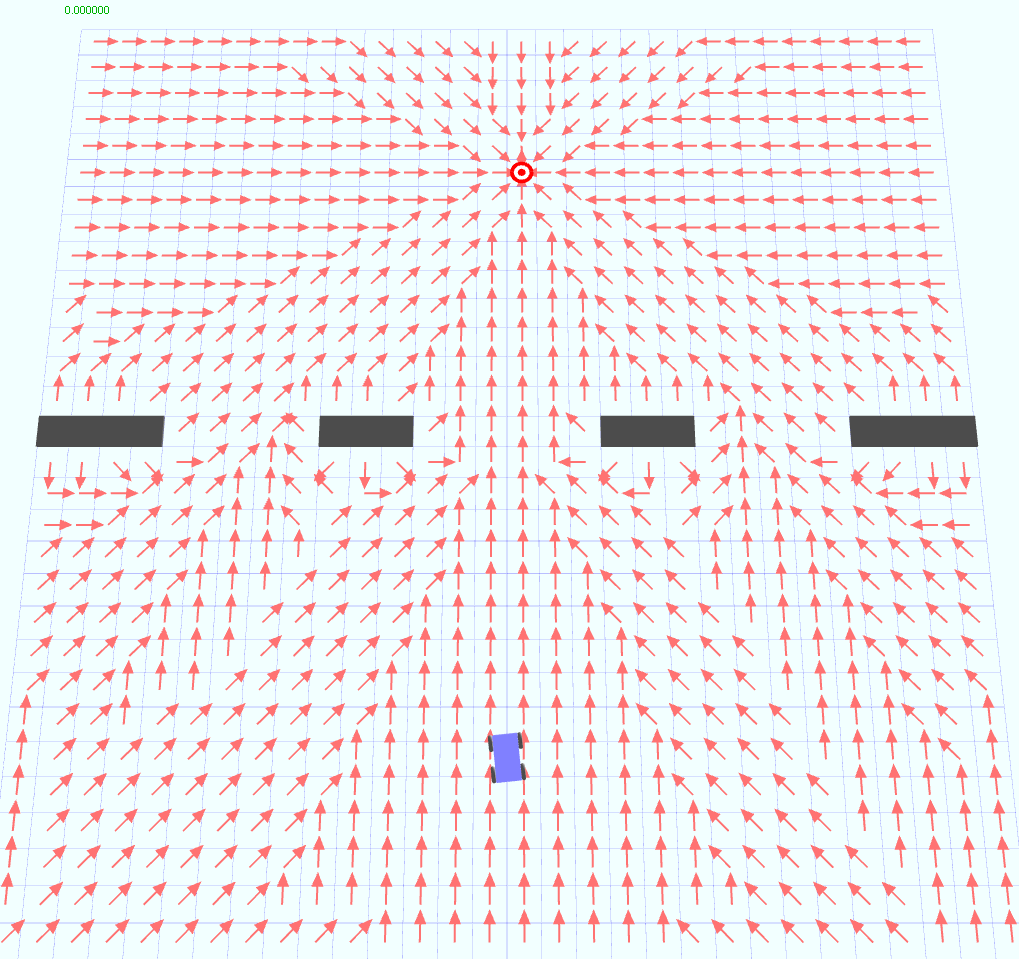}}
	\vspace{-5pt}
	\caption{\small Illustration of reachability landscape. (a) Demonstration of a simple simulation scenario with dark blocks as obstacles, and the goal state as a red circle; (b)-(e) MFPT based reachability landscapes; (f) Converged optimal policy shown in red arrows.}
	\label{fig:heatmap}
	\vspace{-8pt}
\end{figure*}

\subsubsection{Policy Iteration}

The policy iteration (PI) is an anytime algorithm because at any iteration there will be a feasible policy (although not necessarily optimal). 
The PI involves the following two steps. The first step also known as \textit{policy evaluation} is responsible for calculating the value $V(s)$ of each state $s$ given some fixed policy $\pi(s)$ until convergence. 
This is followed by a step called \textit{policy improvement}. Here,
the policy $\pi(s)$ is updated based on the resulting converged values of each state. 
The optimal policy is obtained at convergence when $\pi_{i+1} = \pi_{i}$, where $i$ is the iteration index. 
Each policy improvement step is based on the following equation: 
\begin{equation}\label{eq:policyp}
\begin{split}
\pi(s) = \argmax_{a\in A}  \sum_{s'\in S} &T_a(s, s') \Big( R_a(s, s') +\gamma V^{\pi}(s')\Big).
\end{split}
\end{equation}
There are two ways of computing the policy evaluation step. 
\begin{enumerate}[1)]
    
\item The first approach is to perform Bellman updates iteratively till convergence given some fixed policy using the below equation.
\begin{equation} \label{eq:mdpp}
V_{i+1}^{\pi}(s) = \sum_{s'\in S} T_{\pi(s)}( s, s') \Big(R_{\pi(s)}(s, s')  + \gamma V_i^{\pi}(s') \Big).
\end{equation}
We represent this version of policy iteration where policy evaluation is performed using Bellman updates as \textit{Policy Iteration (PI)} throughout the paper. 

\item The second approach to perform policy evaluation is through linear equation approximation where we solve a linear system as shown in Eq.~\eqref{eq:mdpl}. This outputs the optimal values of each state corresponding to a given policy. 
\begin{equation} \label{eq:mdpl}
\forall{s}, \ \ \ V^{\pi}(s) = \sum_{s'\in S} T_{\pi(s)}( s, s') \Big(R_{\pi(s)}(s, s')  + \gamma V^{\pi}(s') \Big).
\end{equation}
The variables in this system of linear equation are $V^{\pi}(s)$, whereas $T, R$ are constants given a fixed policy. 
We call this version of policy iteration 
as \textit{Policy Iteration - Linear Equation (PI-LE)} throughout the paper.

\end{enumerate}


\vspace{-10pt}
\subsection{Mean First Passage Times}
The {\em first passage time (FPT)}, $T_{ij}$, is defined as the number of state transitions involved in reaching states $s_j$ when started in state $s_i$ for the first time. The {\em mean first passage time (MFPT)}, $\mu_{ij}$ from state $s_i$ to $s_j$ is the expected number of hopping steps for reaching state $s_j$ given initially it was in state $s_i$ ~\cite{AssafSharedShanthikumar1985}. 
The MFPT analysis is built on the Markov chain, and has nothing to do with the agent's actions. 
Remember that, when a MDP is associated to a fixed policy, it then behaves as a Markov chain~\cite{kemeny1959finite}. 

Formally, let us define a Markov chain with $n$ states and transition probability matrix, $p \in {\rm I\!R}^{n,n}$. If the transition probability matrix is regular, then each MFPT, $\mu_{ij} = E(T_{ij})$, satisfies the below conditional expectation formula:
\begin{equation}\label{eq:fpt00}
E(T_{ij}) = \sum_{k} E(T_{ij} | B_k) p_{ik}
\end{equation} 
where, $p_{ik}$ represents the transition probability from state $s_i$ to $s_k$, and $B_k$ is an event where the first state transition happens from state $s_i$ to $s_k$. 
From the definition of mean first passage times, we have, $E(T_{ij} | B_k) = 1 + E(T_{kj})$. So, we can rewrite Eq.~\eqref{eq:fpt00} as follows.
\begin{equation}\label{eq:fpt01}
E(T_{ij}) = \sum_{k} p_{ik} + \sum_{k \neq j} E(T_{kj}) p_{ik}
\end{equation}
Since, $\sum_{k} p_{ik}$ = 1, Eq.~\eqref{eq:fpt01} can be formulated as per the below equation:
\begin{equation}\label{eq:fpt}
\mu_{ij} = 1 + \sum_{k \neq j} p_{ik} \cdot \mu_{kj} \quad \Rightarrow \quad \sum_{k \neq j} p_{ik} \cdot \mu_{kj} - \mu_{ij} = -1,
\end{equation}
Solving all MFPT variables can be viewed as solving a system of linear equations 
\begin{equation}
\label{eq:fpt2}
\begin{spmatrix}{}
    p_{11} - 1 & p_{12} & .. & .. & p_{1n} \\        
    p_{21} & p_{22} - 1 & .. & .. & p_{2n} \\
    .. & .. & .. & .. & .. \\
    .. & .. & .. & .. & .. \\
    p_{n1} & p_{n2} & .. & .. & p_{nn} - 1 \\
\end{spmatrix}
\begin{spmatrix}{}
    \mu_{1j} \\        \mu_{2j} \\ .. \\ .. \\ \mu_{nj}
\end{spmatrix}
=
\begin{spmatrix}{}
    -1  \\        -1 \\ .. \\ .. \\ -1
\end{spmatrix}.
\end{equation}
The values $\mu_{1j}$, $\mu_{2j}$, $....$, $\mu_{nj}$ represents the MFPTs calculated for state transitions from states $s_1$, $s_2$, $....$, $s_n$ to $s_j$ and $\mu_{jj} = 0$. 
To solve above equation, efficient decomposition methods~\cite{Golub1996} may help to avoid a direct matrix inversion.

\vspace{-2pt}
\section{Technical Approach}
\label{technical}


\subsection{Reachability Characterization using MFPT}
The notion of MFPT allows us to define the {\em reachability} of a state.
By ``reachability of a state" we mean that based on current fixed policy, how hard it is for the agent to transit from the current state to the given goal/absorbing state. 
With all MFPT values $\mu_{ij}$ obtained, we can construct a {\em reachability landscape} which is essentially a ``map" measuring the {\em degree of difficulty} of all states transiting to the goal.

Fig.~\ref{fig:heatmap} shows a series of landscapes represented in heatmap in our simulated environment. 
The values in the heatmap range from 0 (cold color) to 100 (warm color). 
In order to better visualize the low MFPT spectrum that we are most interested, any value greater than 100 has been clipped to 100. Fig.~\ref{fig:heatmap0}-\ref{fig:heatmap3} show the change of landscapes as the MFPT-PI algorithm proceeds. 
Initially, all states except the goal state are initialized as unreachable, as shown by the high MFPT color in Fig.~\ref{fig:heatmap0}. 

We observe that the reachability landscape conveys very useful information on potential impacts of different states. 
More specifically, \textbf{a state with a better reachability (smaller MFPT value) is more likely to make a larger change during the MDP convergence procedure, leading to a bigger convergence step.}
With such observation and the new metric, we can design state prioritization schemes for the VI and policy update heuristics for the PI, which are discussed next.


\vspace{-5pt}
\begin{algorithm}[t]
    \caption{Mean First Passage Time based Value Iteration (MFPT-VI)}
    \label{algo:fptvi}
    {\small
        Given states $S$, actions $A$, transition probability $T_{a}(s,s')$ 
        and reward $R_{a}(s,s')$. Assume goal state $s^*$,
        calculate the optimal policy $\pi$\\
                
        \While{true}{
            $V = V'$ \\
            
            
            Calculate MFPT values $\mu_{1s^*}$, $\mu_{2s^*}$, $\cdots$, $\mu_{|S|s^*}$ by solving the linear system as shown in Eq. ~\eqref{eq:fpt2} \\
            
            List $L$ := Sorted states with increasing order of MFPT values \\
            
            \ForEach{state $s$ in $L$} {
                Compute value update at each state $s$ given policy $\pi$:  
                $V'(s) = \max_{a\in A} \sum_{\forall{s'\in S}} T_a( s, s') \Big(R_a(s, s')  + \gamma V(s') \Big)$
            }
                        
            \If{$\max_{s_i} |V(s_i) - V'(s_i)| \leq \epsilon$}{
                break \\ 
            }
          }
        }
\end{algorithm}

\subsection{Mean First Passage Time based Value Iteration (MFPT-VI)}
Classic VI converges the solution through Bellman update as shown in Eq.~\eqref{eq:mdp1}, which essentially sweeps all states by treating them equally important.
The ``prioritized sweeping" mechanism has further improved the convergence rate by exploiting and ranking the states based on their potential contributions, where the metric is based on comparing the difference of values for each state between two consecutive iterations.

The new MFPT-VI method is also built on the prioritized sweeping mechanism, but we propose a metric using the reachability (MFPT values), because as aforementioned, the reachability characterization of each state reflects the potential impact/contribution of this state, and thus provides a natural basis for prioritization. 
Our reachability based metric is different from the existing value-difference metric as follows:
\begin{adjustwidth}{0.7cm}{}
\textit{The reachability landscape can very well capture the degree of importance for all states from a global viewpoint, whereas the classic value-difference strategy evaluates states locally, and may fail grasping the correct global ``convergence direction" due to the local viewpoint.}
\end{adjustwidth}
Note that, since the MFPT computation is relatively expensive, it is not necessary to compute the MFPT at every iteration, but rather after every few iterations.
The computational process of MFPT-VI is pseudo-coded in Alg.~\ref{algo:fptvi}.



\begin{algorithm}[ht]
    \caption{Mean First Passage Time based Policy Iteration (MFPT-PI)}
    \label{alg:fptpi}
    {\small
        Given states $S$, actions $A$, transition probability $T_{a}(s,s')$ and reward
        $R_{a}(s,s')$. Assume goal state $s^*$, calculate the optimal policy $\pi$\\
        Initialize with an arbitrary policy $\pi'$ \\
                
        \While{true}{
            $\pi = \pi'$ \\
            \ForEach{state $s$ of $|S|$} {
                Compute value of each state $s$, V(s) given the policy $\pi$ using Eq.~\eqref{eq:mdpp}
            }
            
            \ForEach{state $s$ of $|S|$} {
                Improve policy $\pi$ at each state $s$ using Eq.~\eqref{eq:policyp}
            }
            
            
            Calculate the MFPT values $\mu_{1s^*}$, $\mu_{2s^*}$, $\cdots$, $\mu_{|S|s^*}$ by solving the linear system in Eq.~\eqref{eq:fpt2}\\
            
            \ForEach{state $s$ of $|S|$} {
                Update policy at state $s$ with the action $a$ that leads to state $s'$ with the minimum MFPT value: $\pi'(s) = \argmin_{a\in A}  \mu_{s's^*}$
            }
            
            \If{$\pi = \pi' $}{
                break \\ 
            }
         }        
    }
\end{algorithm}

\subsection{Mean First Passage Time based Policy Iteration (MFPT-PI)}


Remember that the classic PI algorithm involves two steps: policy evaluation and policy improvement. 
In this sub-section,  we propose a new policy improvement mechanism.  

\textbf{Policy Evaluation:}
Classic policy evaluation utilizes either local policy optimization  (shown in Eq.~\eqref{eq:mdpp}) or a linear equation approximation (shown in  Eq.~\eqref{eq:mdpl}) to obtain a new set of values $V(s),\forall{s\in S}$. 
Our MFPT-PI also employs the local optimization rule described in Eq.~\eqref{eq:mdpp}.

\textbf{Policy Improvement: }
The policy improvement of MFPT-PI is based on MFPT values.
To approximate a good and feasible policy, the operation can be even simpler than the classic approach.
We propose an intuitively simple and computationally cheap heuristic which shows great performance: 
the action selection can be following (or weighted by) the ``gradient" of reachability landscape. 
For instance, in the agent motion planning scenario where each state transits to some set of states in the vicinity, the locally optimal action can be selected to transit to the neighbouring state with the least MFPT value. 

Alternating the above two steps allow us to converge to the solution.
Fig.~\ref{fig:heatmap} illustrates the basic idea. In the beginning, a random policy is initialized and all states except the goal are pre-set with high MFPT values, see Fig.~\ref{fig:heatmap0}.
In the first iteration, after completing the policy evaluation, the reachability landscape is shown as Fig.~\ref{fig:heatmap1}, which  characterizes the reachability regions at a high level. The result is more like an image segmentation solution where regions with distinct reachability levels are roughly partitioned: since in the middle area there are a few obstacles that separate the space, the reachability landscape is split into two regions, with upper area being well reachable and the lower part fully un-reachable.
With a couple more iterations, the reachability landscape is refined until low level details are well characterized (see Figs~\ref{fig:heatmap2}-\ref{fig:heatmap3}). 
In essence, a major benefit of the proposed MFPT-PI algorithm is that: 
at earlier stage of the convergence, it captures the big picture; and later it focuses on improving the details once the policy becomes better. 

It is worth mentioning that, if there are multiple goal/absorbing states, then each goal state will require to compute its own MFPT landscape, and the final reachability landscape will be normalized by all MFPTs corresponding to all terminal states.




\subsection{Time Complexity Analysis}

The classic VI algorithm has a time complexity of $O(|A||S|^2)$ per iteration where $|S|$ represents number of states and $|A|$ represents number of actions. The classic PI has a time complexity of $O(|A||S|^2 + |S|^{2.3})$ if a linear equation solver is used (the matrix decomposition has a time complexity of $O(|S|^{2.3})$ if state-of-the-art algorithms are employed~\cite{Golub1996}, given that the size of matrix is the number of states $|S|$). 
Since, both the proposed MFPT-VI and MFPT-PI algorithms involve a key step for calculating the MFPT which also needs to solve a linear system, 
therefore, for each iteration, both the MFPT-VI and MFPT-PI algorithms have a time complexity of $O(|A||S|^2 + |S|^{2.3})$.
Although, the MFPT is an extra component (and cost) introduced to the VI and PI mechanisms, our reachability characterization enables the MDP solving process to converge in much fewer steps. 
An important point to note here is that the MFPT values don't need to be computed frequently at every iteration and it is only calculated when characterizing global feature is needed, which also reduces the practical runtime. 

It is also worth mentioning that, although the time complexity for matrix decomposition is $O(|S|^{2.3})$ in general, 
for sparse matrices, efficient decomposition techniques perform much faster than $O(|S|^{2.3})$ where the time complexity depends on the number of non-zero elements in the matrix. For example, in certain robotic motion planning scenario, the robot transits to states within a certain vicinity causing the transition probability matrix to be sparse by nature.
Inspite of this additional time complexity per iteration, MFPT based algorithms have faster overall runtime than existing solution methods because they converge in much fewer steps. 
Results are presented in the following section.



\section{Experimental Results}
\label{Results}


\subsection{Experimental Setting}
\subsubsection*{\textbf{Task Details}}
We validated our method through numerical evaluations with two different simulators running on a Linux machine.
We consider the MDP problem where each action can lead to transitions to all other states with certain transition probabilities.
However, in many practical scenarios, the probability of transiting from a state to another state that is ``weakly connected" can be small, even close to 0. This can potentially result in non-dense transition matrix.
For example, in the robotic motion planning scenario, the state transition probability from state $s_i$ to state $s_j$ can be correlated with the time or distance of traveling from $s_i$ to $s_j$, and it is more likely for a state to transit to some states within certain vicinity.  

\begin{figure}[t]
  \centering
  \subfigure[]
        {\label{fig:S_Env}\includegraphics[height=1.3in]{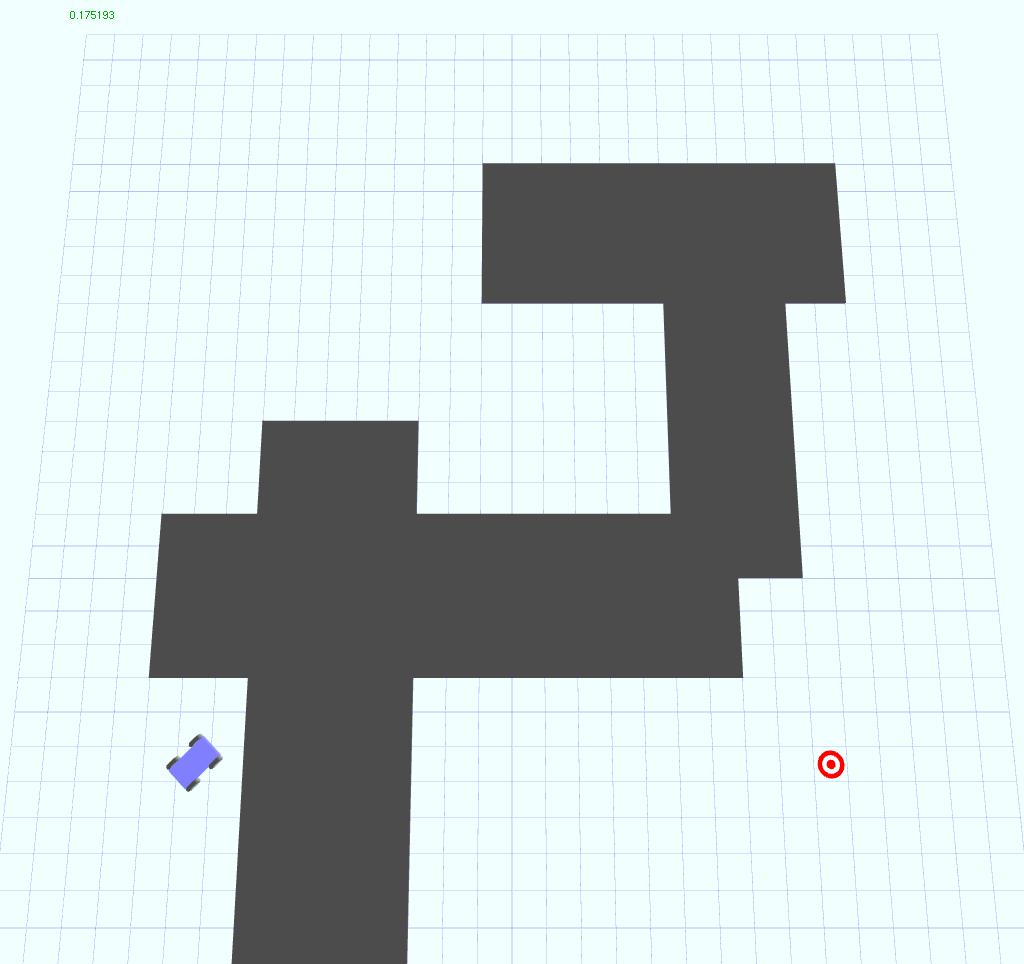}}
        \quad \quad 
  \subfigure[]
        {\label{fig:trajectory}\includegraphics[height=1.3in]{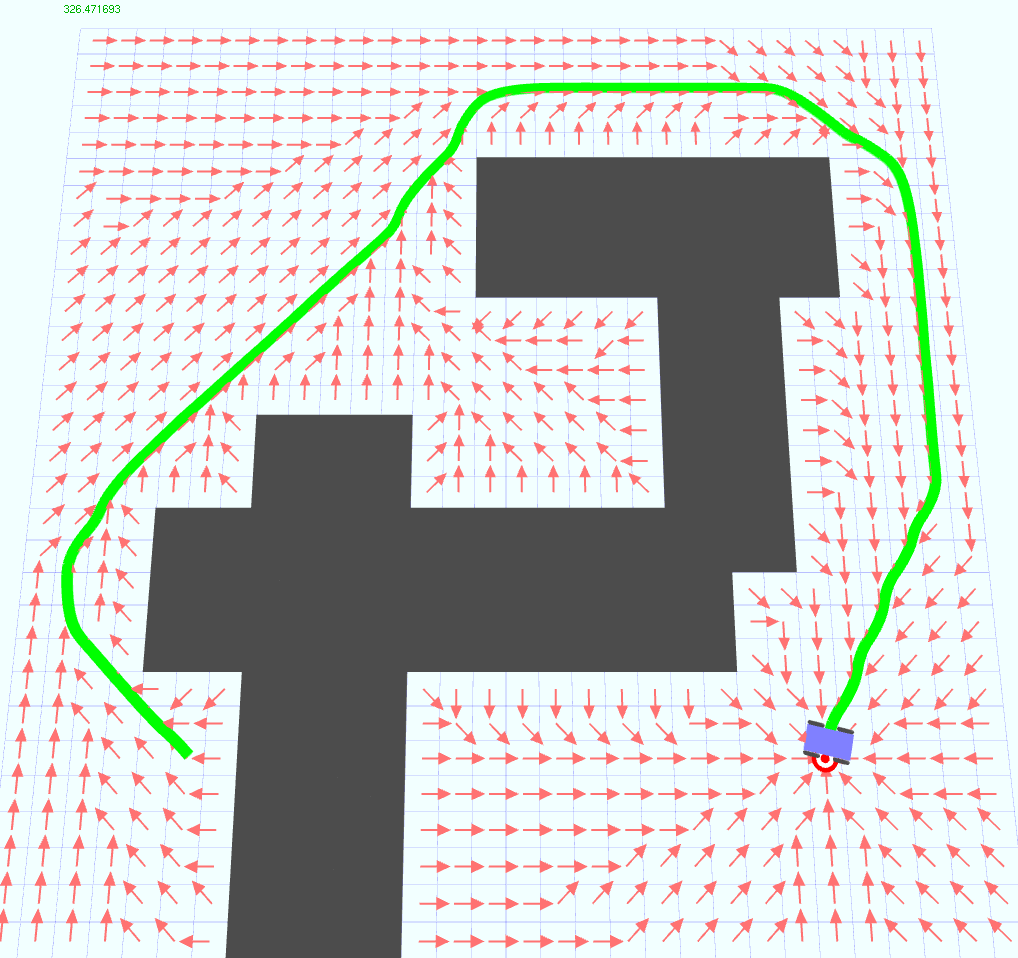}} \\
  \subfigure[]
        {\label{fig:heatmap01}\includegraphics[height=0.6in]{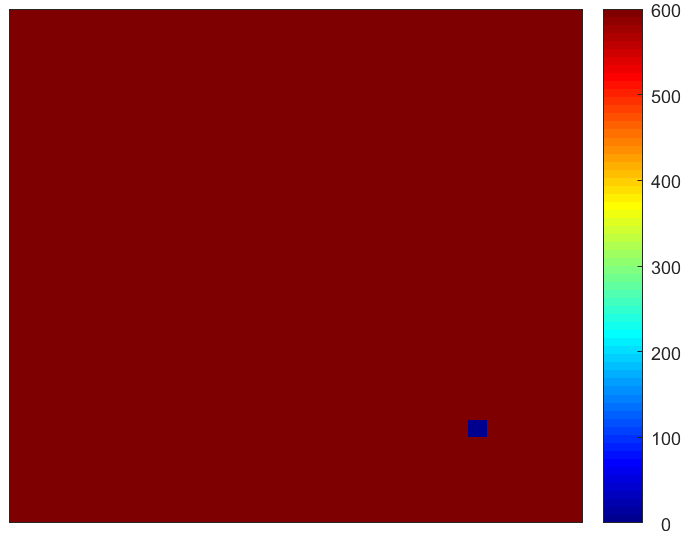}}
  \subfigure[]    
        {\label{fig:heatmap11}\includegraphics[height=0.6in]{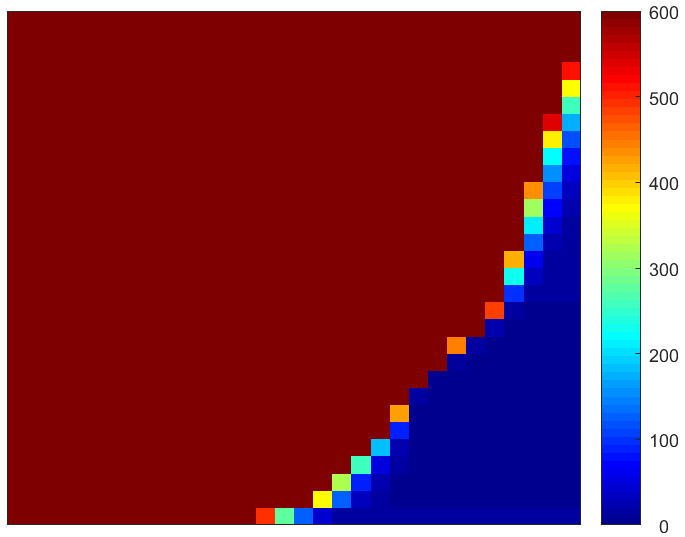}} 
  \subfigure[]
        {\label{fig:heatmap21}\includegraphics[height=0.6in]{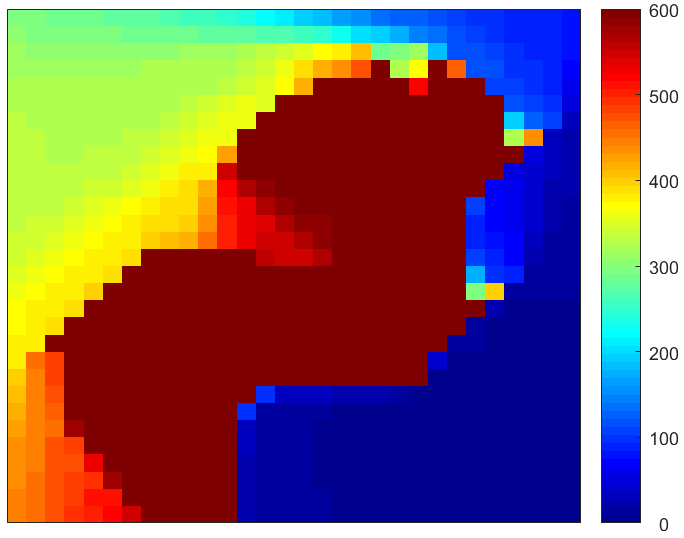}}  
  \subfigure[]
        {\label{fig:heatmap41}\includegraphics[height=0.6in]{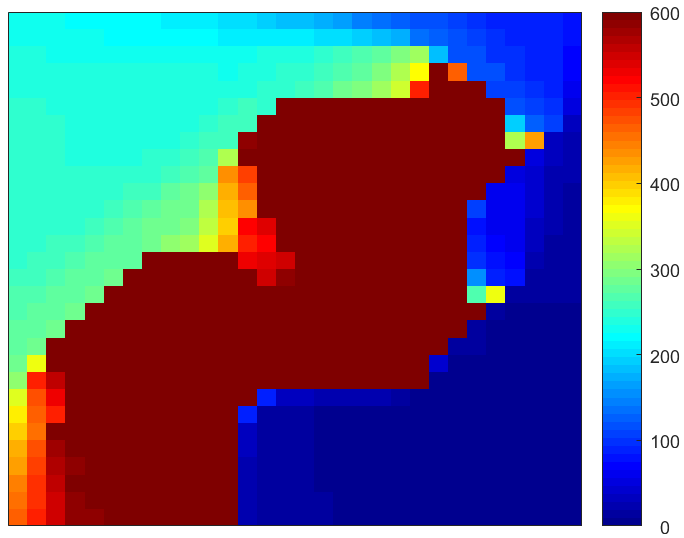}}

	\vspace{-7pt}
	\caption{\small (a) Demonstration of simulation environment, with the agent's initial state (blue) and the goal state (red). Grey blocks are obstacles; (b) Converged optimal policy (red arrows) and a trajectory completed by the agent to reach the goal; (c)-(f) Evolution of  reachability landscapes.}
	\label{fig:heatmapApp}
	\vspace{-10pt}
\end{figure}

For the first task, we developed a simulator in C++ using OpenGl. To obtain the discrete MDP states, we tessellate the agent's workspace into a 2D grid map, and represent the center of each grid as a state.
In this way, the state hopping between two states represents the corresponding motion in the workspace. Each non-boundary state has a total of nine actions, i.e., a non-boundary state can move in the directions of N, NE, E, SE, S, SW, W, NW, plus an idle action returning to itself.
A demonstration is shown in Fig.~\ref{fig:heatmapApp}.

For the second task, we developed a simulator in C++ using ROS and rviz. The agent's workspace is partitioned into a 3D grid map where the center of each grid represents a MDP state. Each non-boundary state has a total of five actions, i.e., a non-boundary state can move in the directions of N, E, S, W, TOP, BOTTOM plus an idle action. 

In both tasks, the objective of the agent is to reach a goal state from an initial start location in the presence of obstacles. The reward function for both setups is represented as high positive reward at the goal state and -1 for obstacle states. All experiments were performed on a laptop computer with 8GB RAM and 2.6 GHz quad-core Intel i7 processor.

\subsection{2D Grid Setup}

In this setup, we compare our proposed algorithms with their corresponding baseline algorithms in terms of the practical runtime performance, iterations required to converge and their convergence profile. Later, we also performed some analysis on time costs of individual components and MFPT invoking frequency.

\begin{figure}[t]
  \centering
  \subfigure[]
        {\label{fig:TS_V}\includegraphics[height=1.3in]{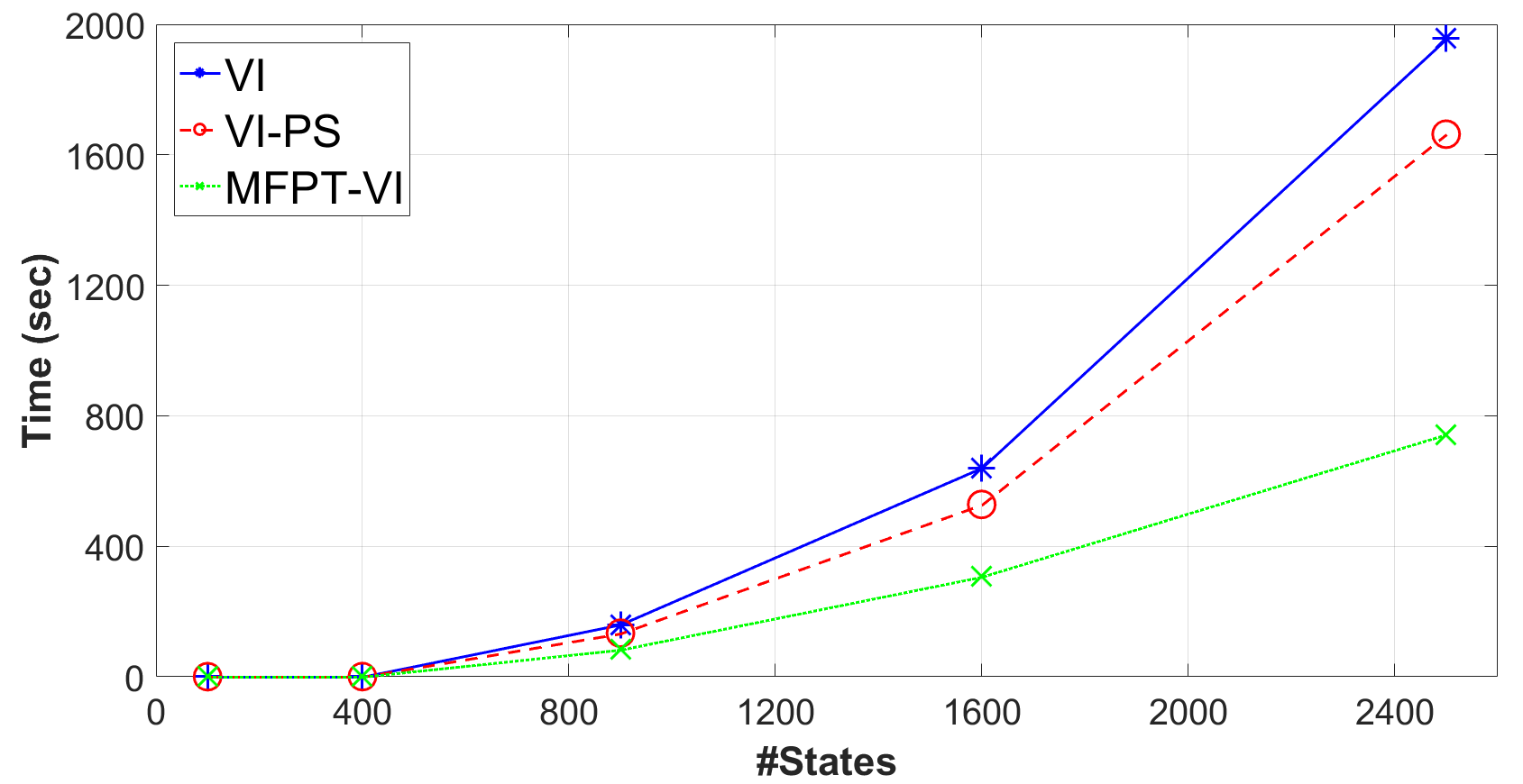}}
        \quad
  \subfigure[]    
        {\label{fig:TS_P}\includegraphics[height=1.3in]{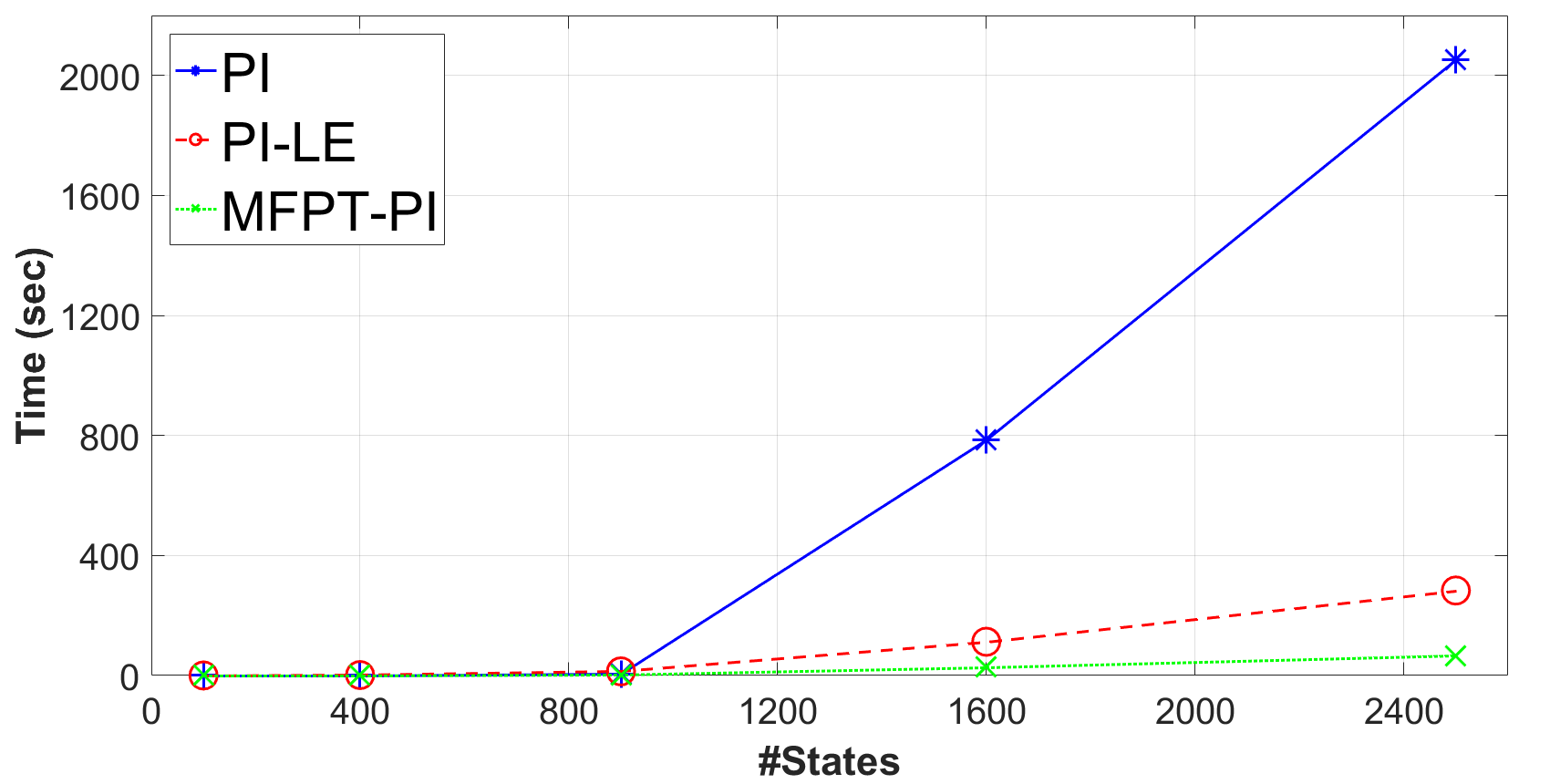}}  
	\vspace{-5pt}
	\caption{\small Time comparisons between the baseline methods and our proposed algorithms, with changing numbers of states ($x$-axis). (a) Variants of value iteration methods. (b) Variants of policy iteration methods.
	}
	\label{fig:TimeVsStates}
	\vspace{-10pt}
\end{figure}

\textbf{Practical Runtime Performance:}
We first investigate the time taken by the algorithms with changing number of states. 
Fig.~\ref{fig:TS_V} compares the time taken by VI, VI-PS and MFPT-VI algorithms. 
In MFPT-VI, the MFPT component is computed every three iterations. Also, the convergence thresholds are set the same for all methods.
The results show that VI is the slowest, and VI-PS is faster than VI due to the prioritized sweeping heuristic. And our proposed MFPT-VI is much faster than the other two algorithms. 
Although, time taken per iteration for MFPT-VI is higher than that of the classic VI, overall the MFPT-VI takes the least time due to much fewer convergence steps and also because MFPT values are computed every three iterations. Below, we will further elaborate on why we have decided not to calculate the MFPT values every iteration.

Fig.~\ref{fig:TS_P} compares the time taken by PI, PI-LE, and MFPT-PI. 
The results also show that the MFPT-PI is the fastest compared to the other two algorithms. 
Note that, both MFPT-PI and PI-LE require to compute linear equations, and in our implementation we use the \textit{Eigen} library. (Since the reachability characterization does not require high MFPT accuracy, among many provided decomposition solvers we chose one of the fast variants which however do not provide the highest accuracy.)

\begin{figure}[t]
  \centering
  \subfigure[]
        {\label{fig:CS_V}\includegraphics[height=1.4in]{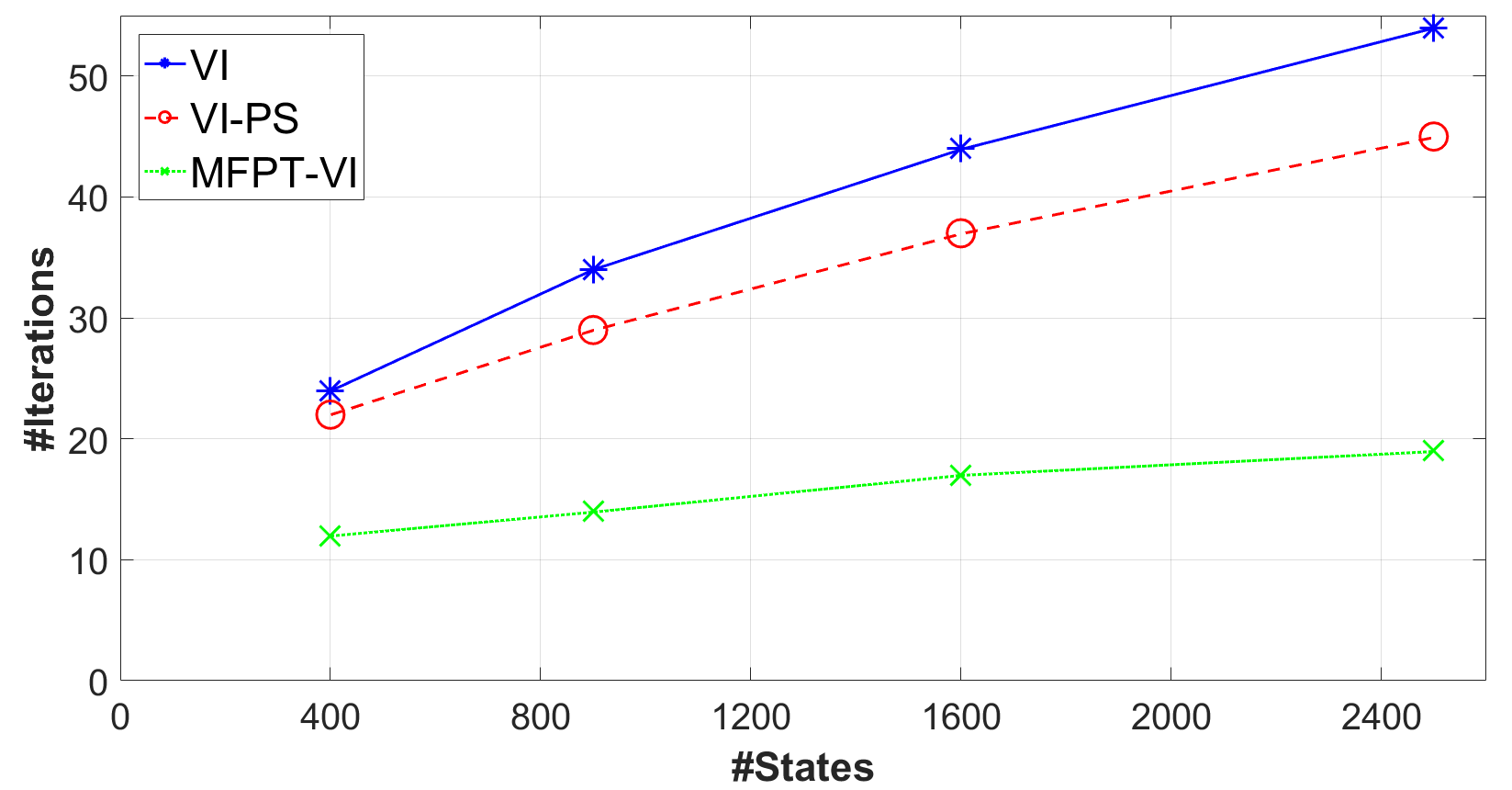}}
        \quad
  \subfigure[]    
        {\label{fig:CS_P}\includegraphics[height=1.4in]{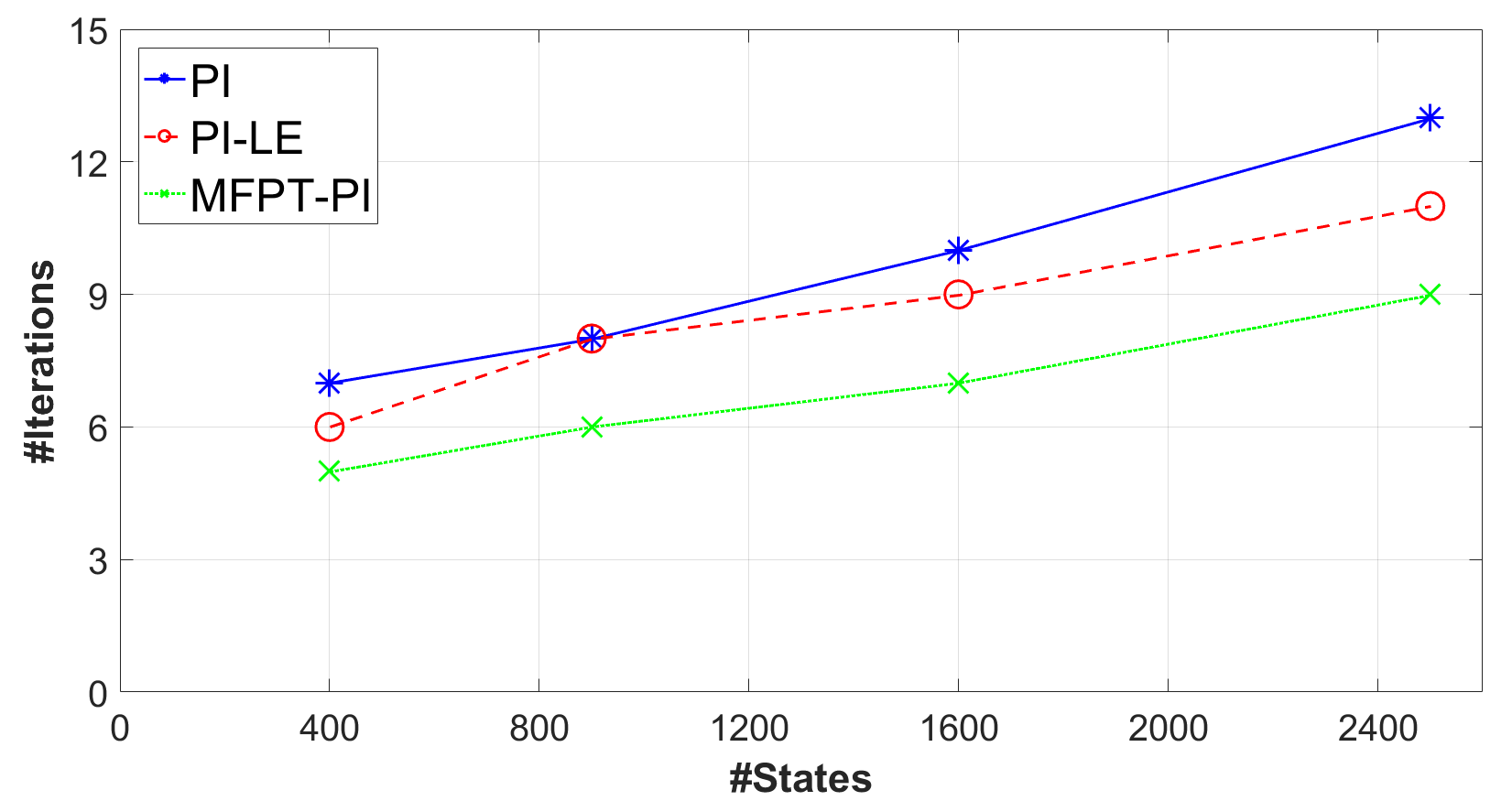}}	
    \vspace{-5pt}
	\caption{\small Iteration to convergence between the baseline methods and our proposed algorithms, with changing numbers of states ($x$-axis). (a) Variants of value iteration methods. (b) Variants of policy iteration methods.
	}
	\label{fig:ConvergenceVsStates}
	\vspace{-10pt}
\end{figure}

\textbf{Number of Iterations: }
We then analyze the number of iterations taken by the algorithms as the number of states change.
Fig.~\ref{fig:CS_V} compares the number of iterations taken by VI, VI-PS and MFPT-VI, respectively. 
The results reveal that MFPT-VI converges the fastest compared to the other two algorithms. 
Fig.~\ref{fig:CS_P} compares the iterations taken by PI, PI-LE, and MFPT-PI. 
Again, we can see that MFPT-PI converges the fastest among all three algorithms.



\begin{figure}[ht]
  \centering
  \subfigure[]
        {\label{fig:C_V}\includegraphics[height=0.82in]{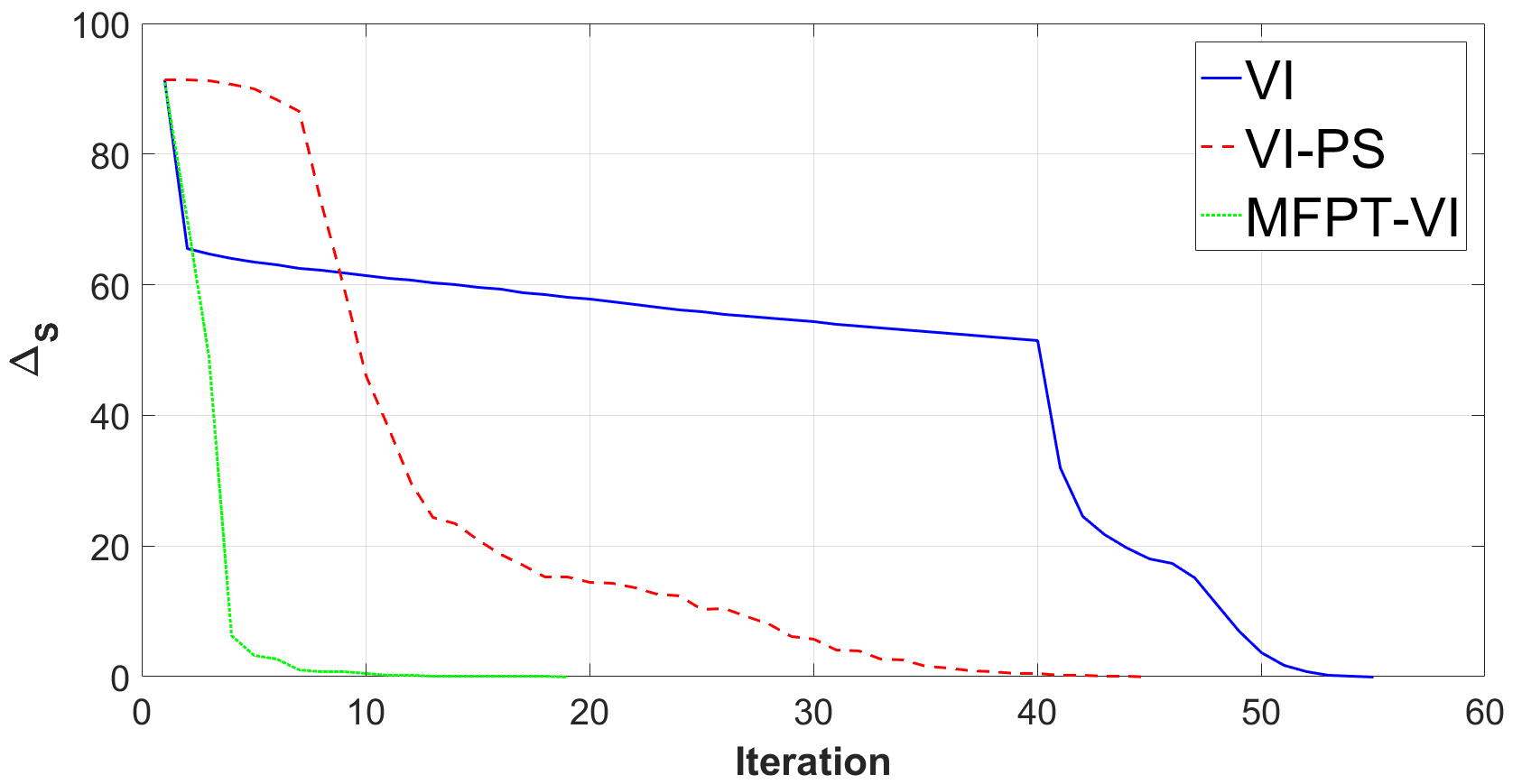}}
        \quad
  \subfigure[]    
        {\label{fig:C_P}\includegraphics[height=0.82in]{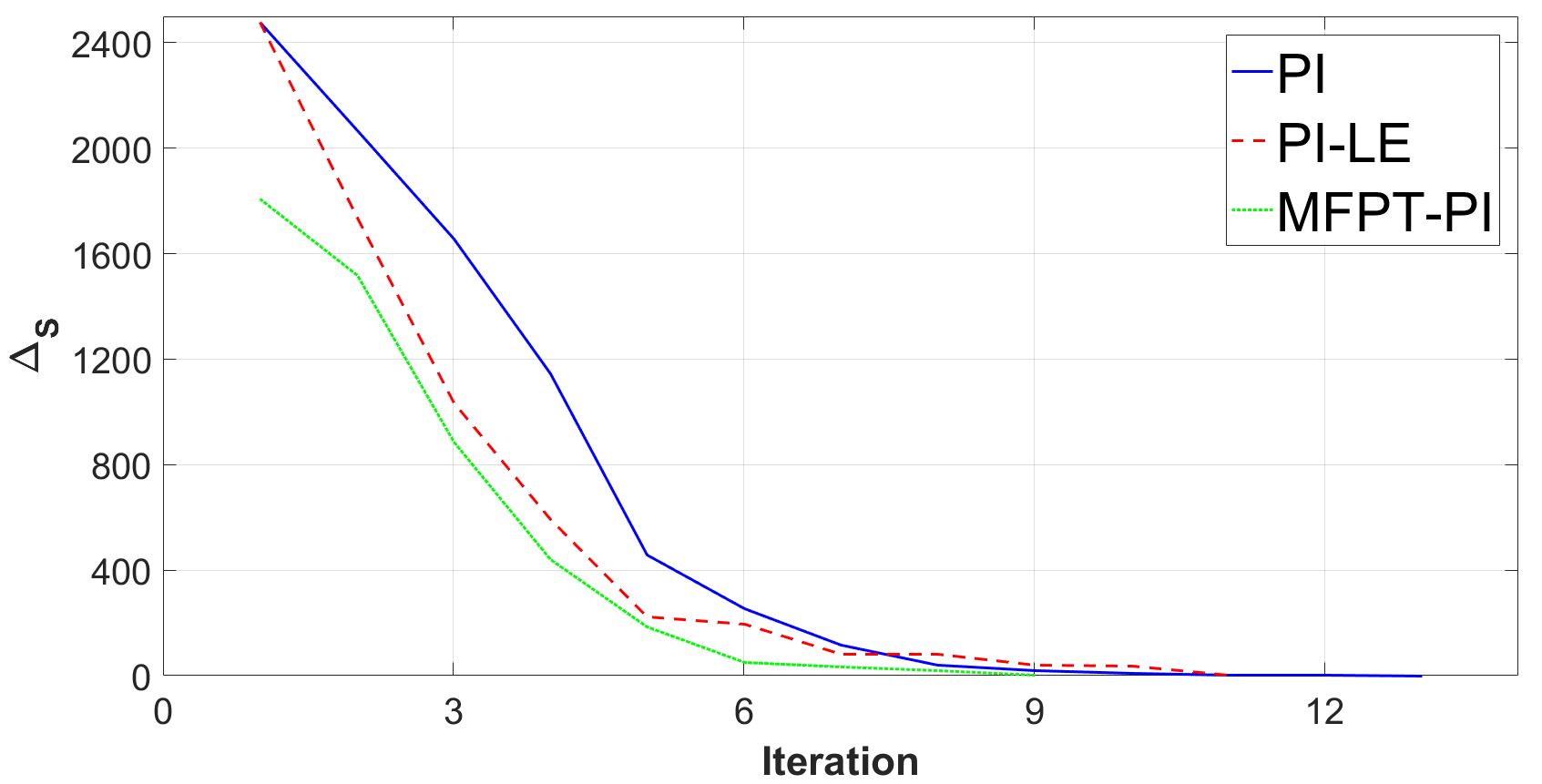}}
	\vspace{-5pt}
	\caption{\small (a) The progress of VI, VI-PS and MFPT-VI across iterations. (b) The progress of PI, PI-LE and MFPT-PI across iterations.}
	\label{fig:progress}
	\vspace{-8pt}
\end{figure}

\textbf{Convergence Profile: }
Here, we investigate the detailed convergence profile along with the increasing number of iterations. 
Since one important criterion to judge the convergence is to see if the error/difference between two consecutive iterations is small enough, thus, we utilize the maximum error $\Delta_S$ across all states as an evaluation metric. 
Specifically, 
for VI, VI-PS, MFPT-VI, the error $\Delta_S$ is defined as 
\begin{equation}\label{eq:delta_vi}
\Delta_S = \max_{s_i \in S} |V(s_i) - V'(s_i)|
\end{equation}
where $V$, $V'$ represent the values of states at iteration $i$ and $i+1$.
For PI, PI-LE and MFPT-PI, the $\Delta_S$ is defined as the total number of mismatches between two consecutive iterations' policies: 
\begin{equation}
    \Delta_S = \sum_{s_i \in S}I_{\pi(s_i)}, \text{ where } I_{\pi(s_i)} = 
    \begin{cases}
     & 0, \text{ if } \pi(s_i)=\pi'(s_i) \\ 
     & 1, \text{ otherwise }
    \end{cases}
\end{equation}
where $\pi$ and $\pi'$ represent the policy at iteration $i$ and $i+1$, respectively.

Fig.~\ref{fig:C_V} shows that initially VI, VI-PS and MFPT-VI start with the same $\Delta_S$ value. 
It is very obvious that our MFPT-VI approach has an extremely steep decrement in the first few iterations. 
For example, in our simulation scenario with around 2500 states and a threshold value of 0.1, the MFPT-VI converged in only 19 iterations;  in contrast, VI took 55 iterations and VI-PS took 45. 

Fig.~\ref{fig:C_P} shows profiles for PI, PI-LE and MFPT-PI. 
Note that, with the policy obtained after the first iteration, MFPT-PI already produced a smaller $\Delta_S$ value compared to the other two algorithms. 
After that, the $\Delta_S$ values in case of MFPT-PI are lower than those of PI and PI-LE.
For instance, in the same simulation example mentioned above, our MFPT-PI method converged in 9 iterations whereas PI took 13 iterations and PI-LE took 11 iterations. The profiles also indicate that the PI based methods take much fewer iterations, but more time per iteration, than those of VI based approaches.


\begin{figure}
  \centering
  \subfigure[]
        {\label{fig:P_VI}\includegraphics[height=0.9in]{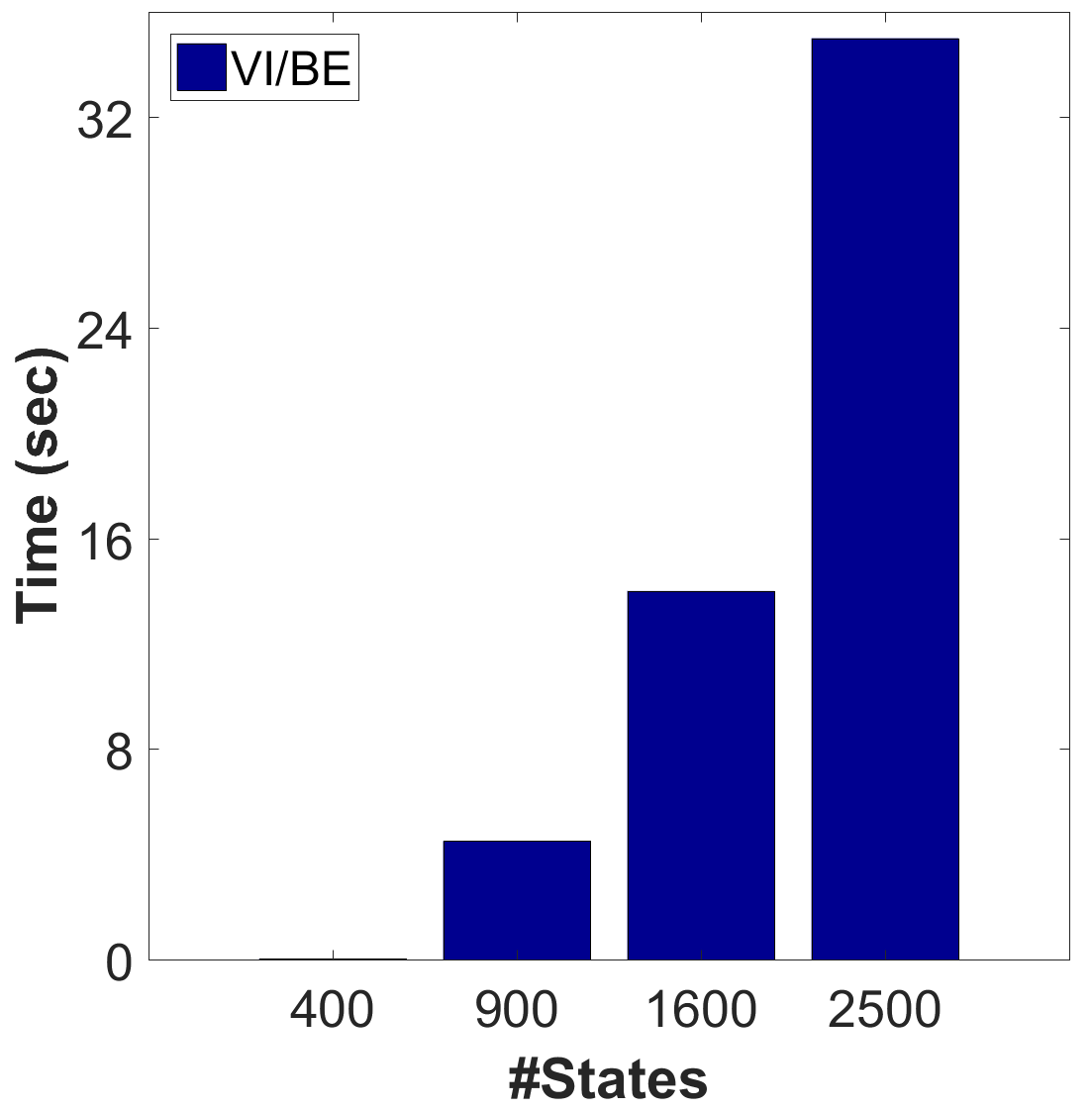}}
        \quad
  \subfigure[]
        {\label{fig:P_VI_PS}\includegraphics[height=0.9in]{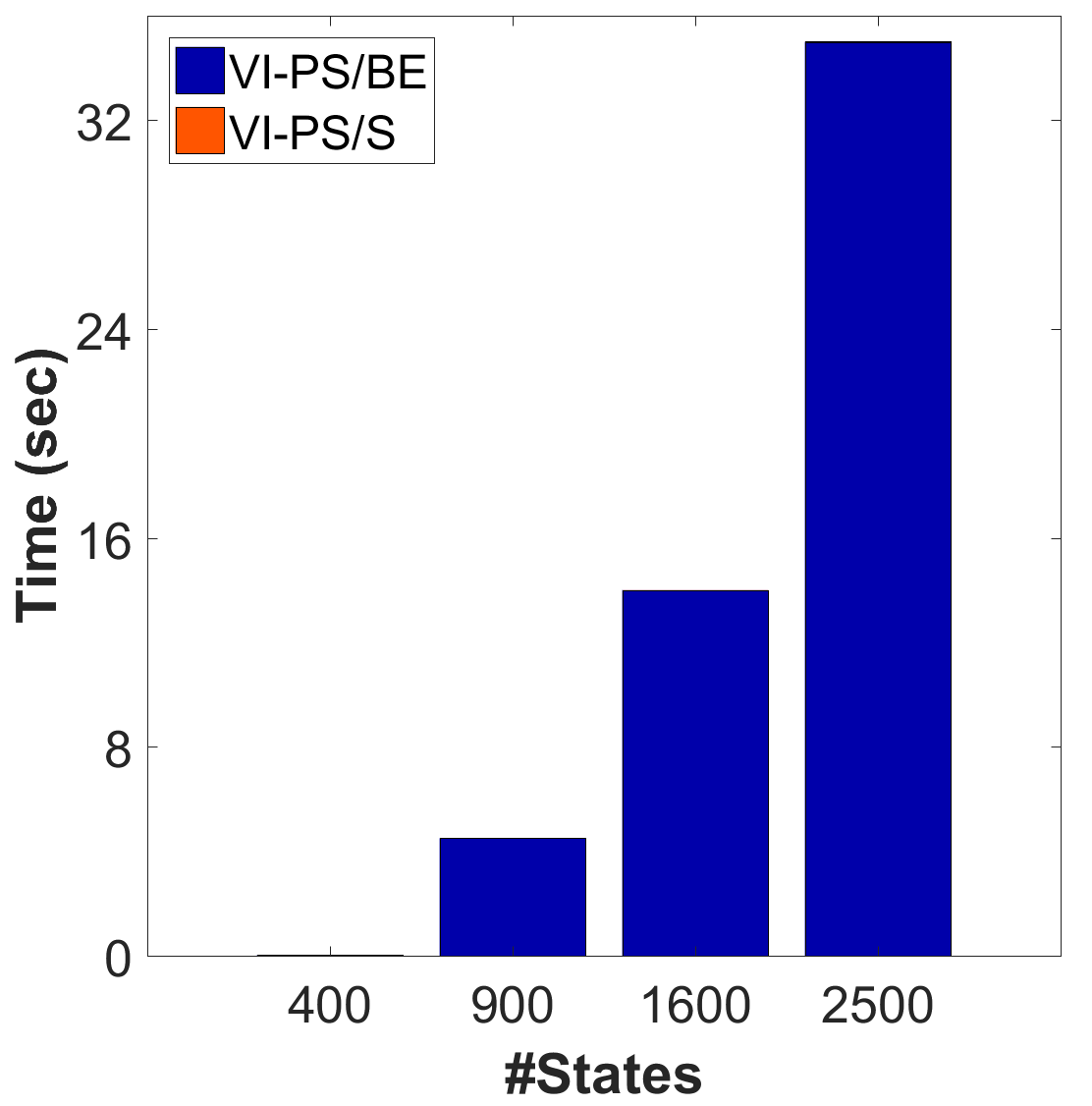}}
         \quad
  \subfigure[]
        {\label{fig:P_VI_FPT}\includegraphics[height=0.9in]{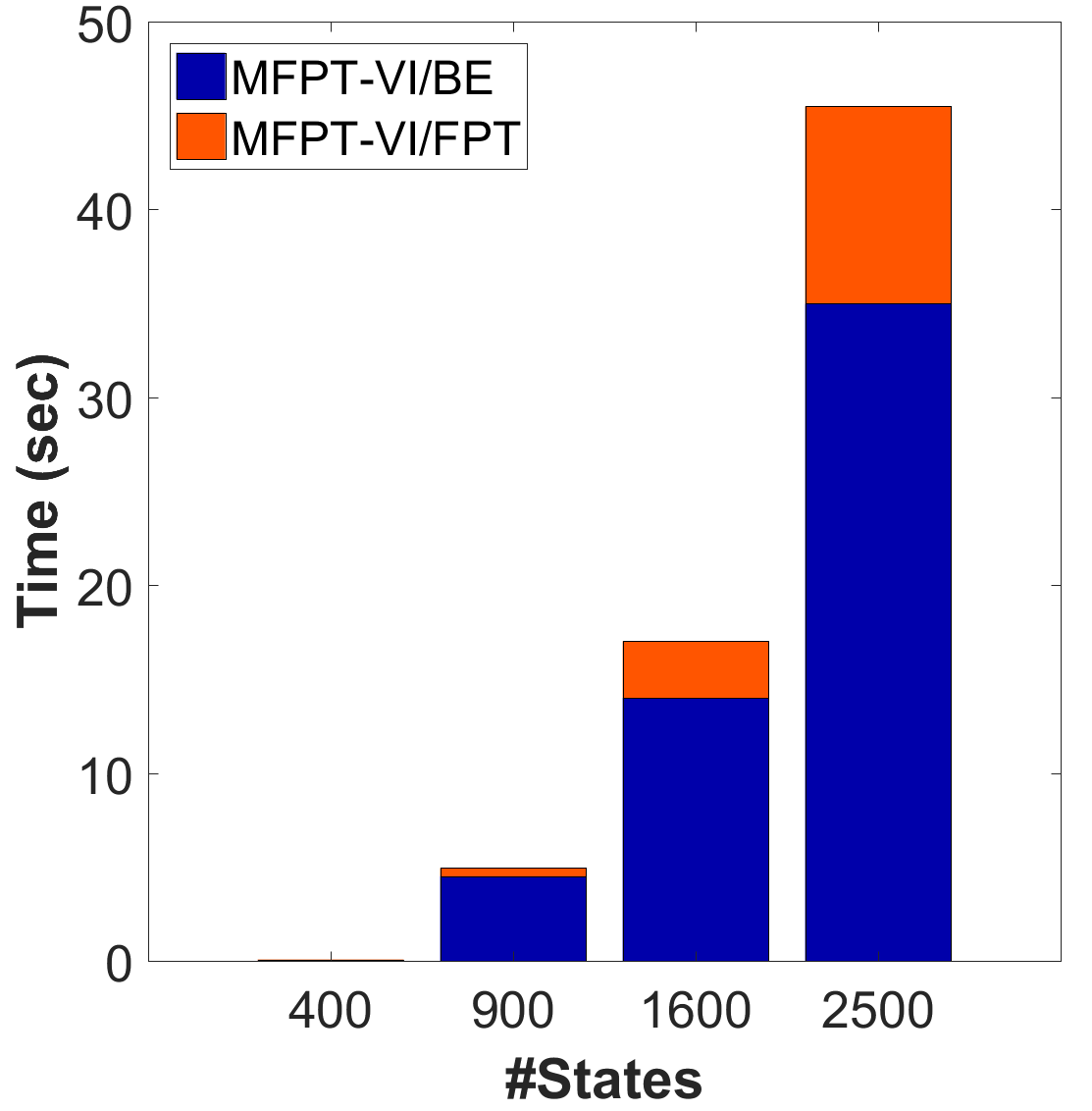}}
    \quad 
  \subfigure[]
        {\label{fig:P_PI}\includegraphics[height=0.9in]{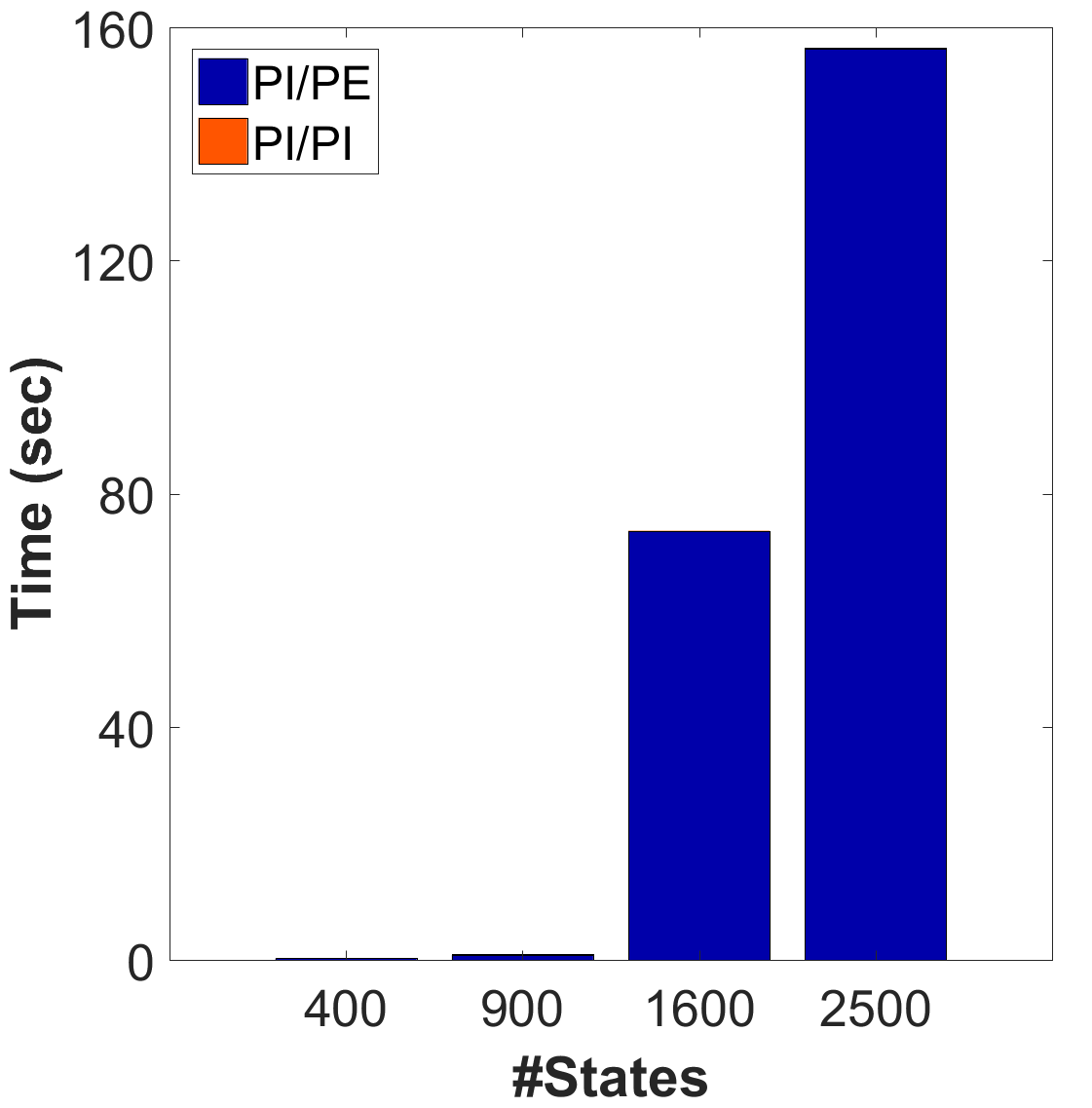}}
        \quad
  \subfigure[]    
        {\label{fig:P_PI_Linear}\includegraphics[height=0.9in]{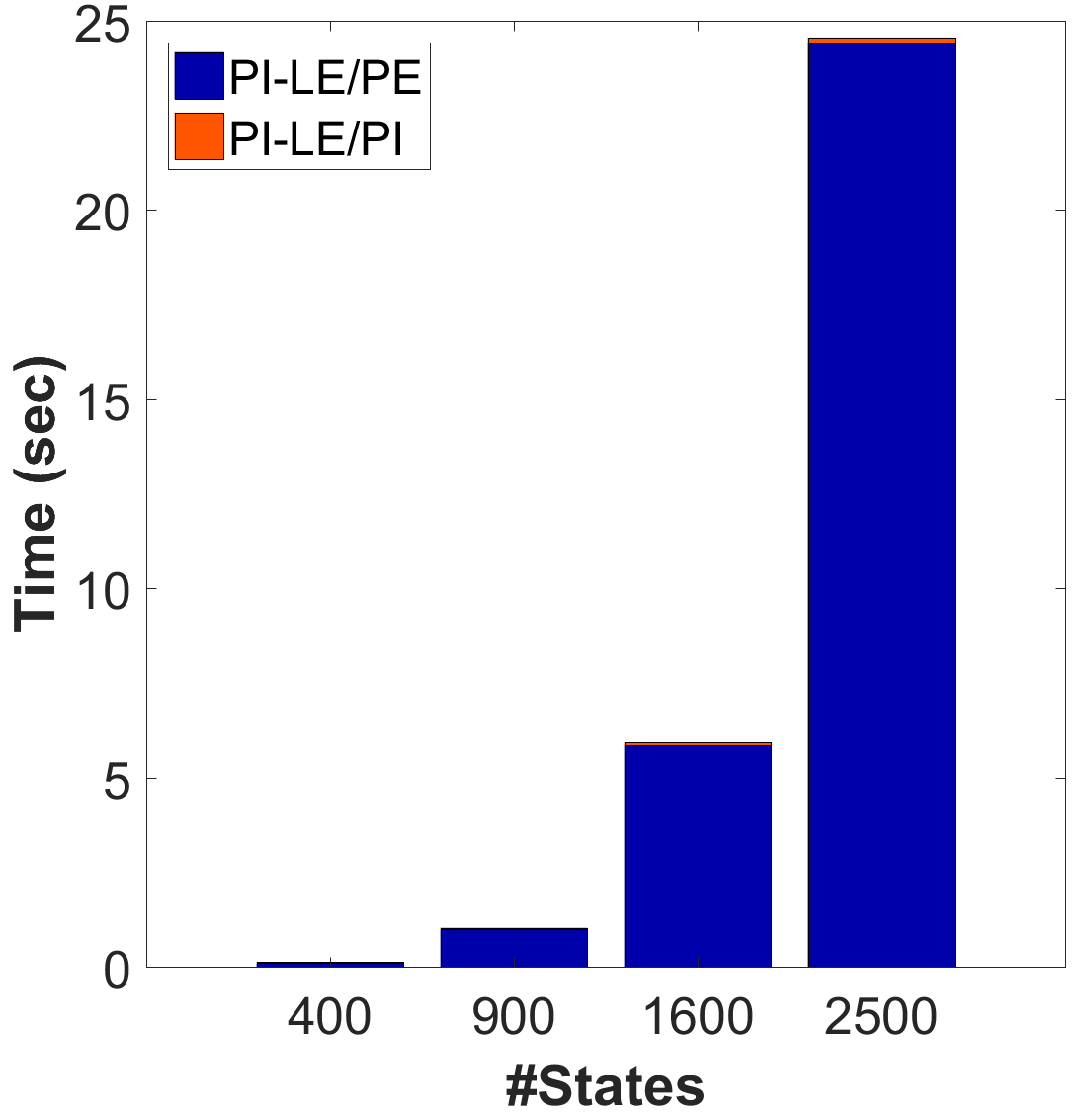}}
        \quad
  \subfigure[]
        {\label{fig:P_PI_FPT}\includegraphics[height=0.9in]{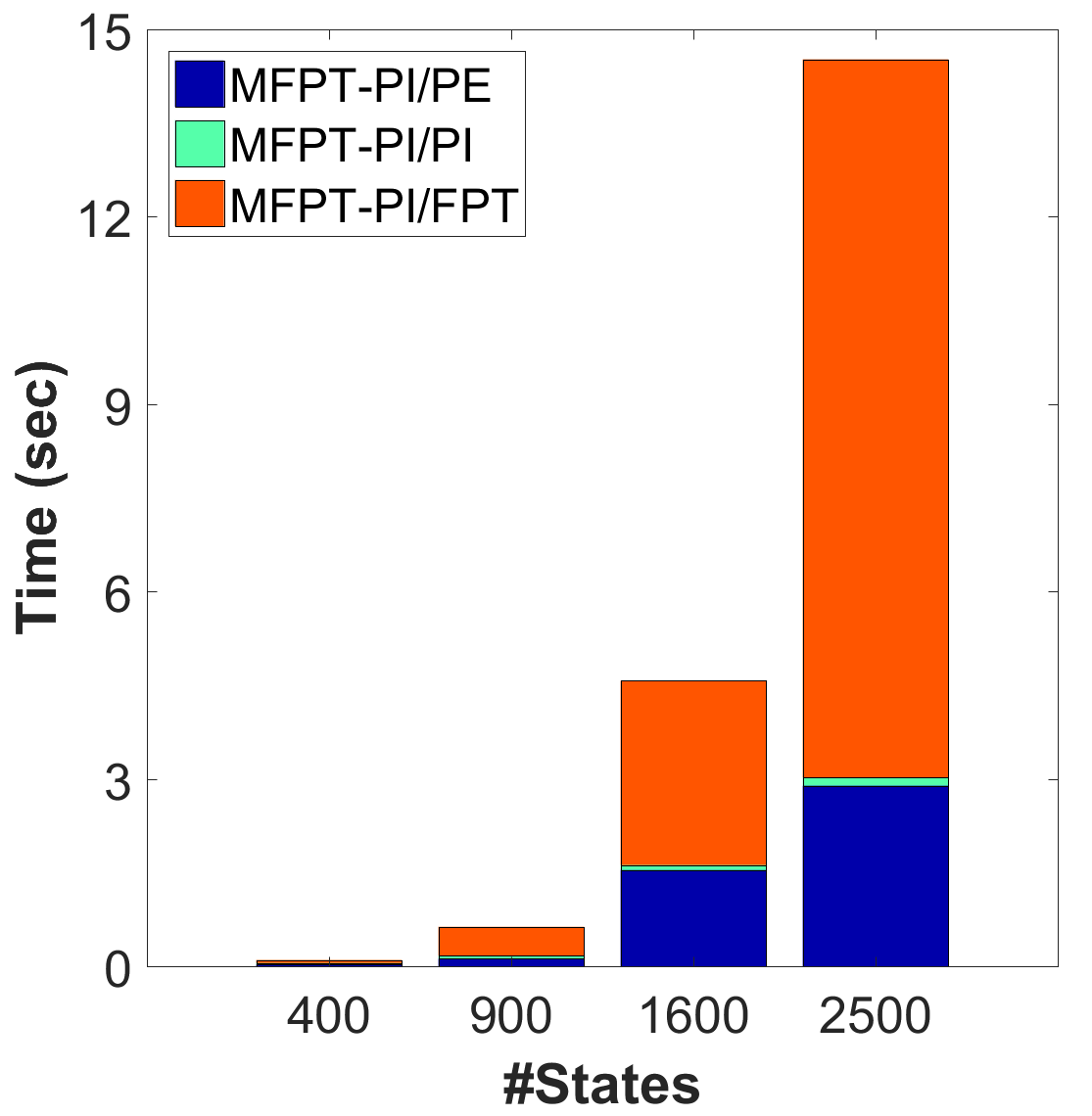}} 
	\vspace{-5pt}
	\caption{\small Time taken by individual components of algorithms. (a)-(c) Variants of value iteration methods. VI/BE, VI-PS/BE, MFPT-VI/BE represents the time involved in computing bellman equation for VI, VI-PS and MFPT-VI respectively whereas MFPT-VI/FPT is the time involved for computing MFPT values. (d)-(f) Variants of policy iteration methods. PI/PE, PI-LE/PE, MFPT-PI/PE represents the time involved in computing the policy evaluation step for PI, PI-LE and MFPT-PI respectively. PI/PI, PI-LE/PI, MFPT-PI/PI represents the time involved in computing the policy improvement step for PI, PI-LE and MFPT-PI respectively. MFPT-PI/FPT is the time involved for computing MFPT values.}
	\vspace{-10pt}
\end{figure}

\textbf{Time Costs of Components: }
We also looked into the detailed time taken by critical components of each algorithm.
The VI related algorithms are analyzed in Fig.~\ref{fig:P_VI}-\ref{fig:P_VI_FPT}. 
Specifically, for VI most time is used for Bellman backup operations as shown in Fig.~\ref{fig:P_VI};
for VI-PS there is an additional step involved in sorting the states for the purpose of states prioritization, which is negligible when compared with the cost of Bellman backup, as illustrated by Fig.~\ref{fig:P_VI_PS}.
In contrast, in our MFPT-VI algorithm, a big chunk of time is used for computing the MFPT values, as shown in Fig.~\ref{fig:P_VI_FPT}.

Fig.~\ref{fig:P_PI}-\ref{fig:P_PI_FPT} show results for PI based algorithms. 
We recorded the time taken by the policy evaluation and policy improvement phases. 
In addition, for MFPT-PI, we also recorded the time used by MFPT calculation. 
From the results we can see that, policy evaluation dominates the time in PI and PI-LE,
whereas for MFPT-PI, the MFPT calculation takes most of the time as shown in Fig.~\ref{fig:P_PI_FPT}. 
However, MFPT-PI compensated for the expensive MFPT calculation by converging in fewer iterations, leading to a faster convergence in general.

\textbf{MFPT Invoking Frequency: }
As mentioned earlier, the MFPT values are computed after every few iterations when there is a need for characterizing global features. Next, we present some analysis related to this.

\begin{figure}
  \centering
  \subfigure[]
        {\label{fig:TimeIter}\includegraphics[height=0.83in]{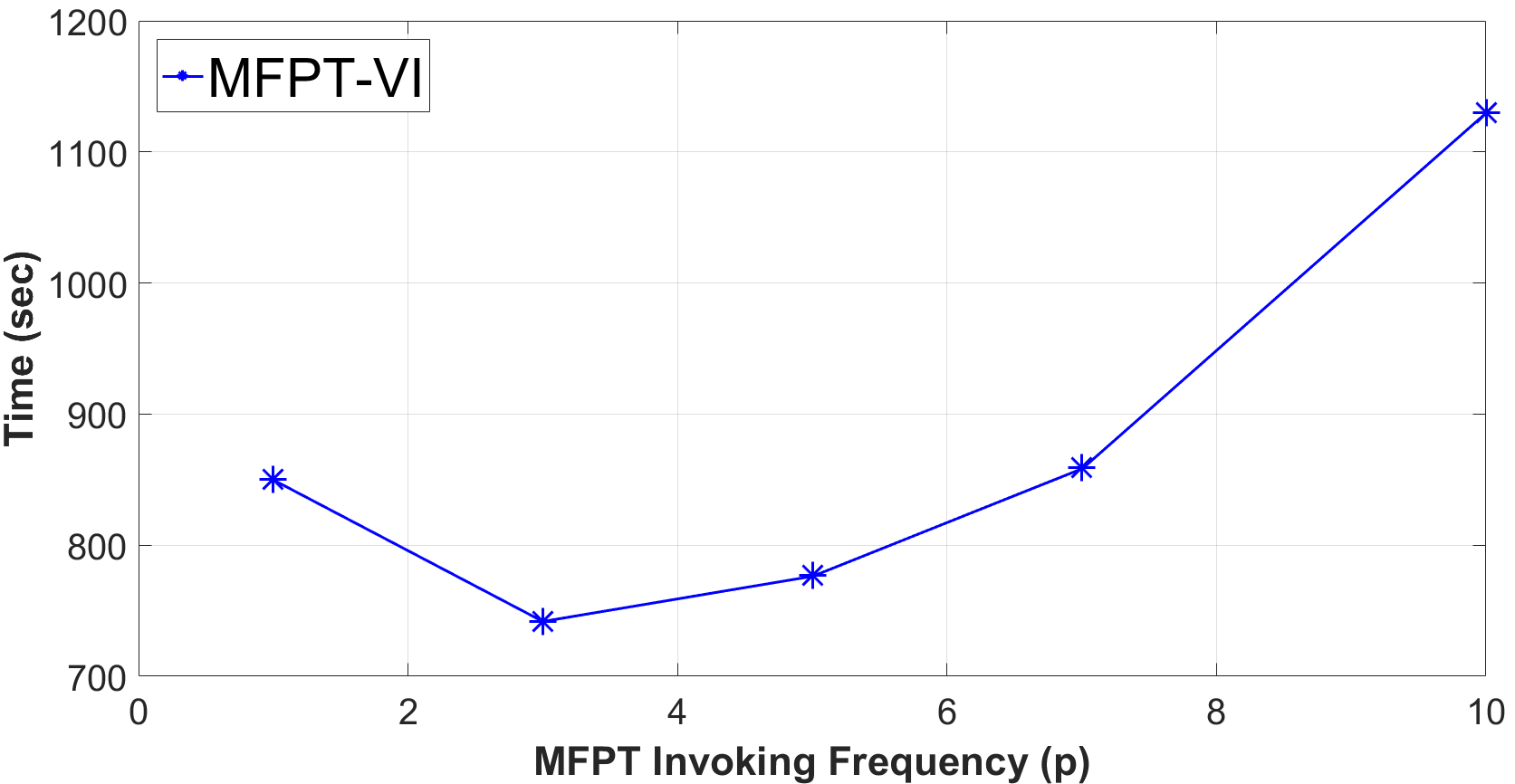}}
        \quad
  \subfigure[]    
        {\label{fig:ConvIter}\includegraphics[height=0.83in]{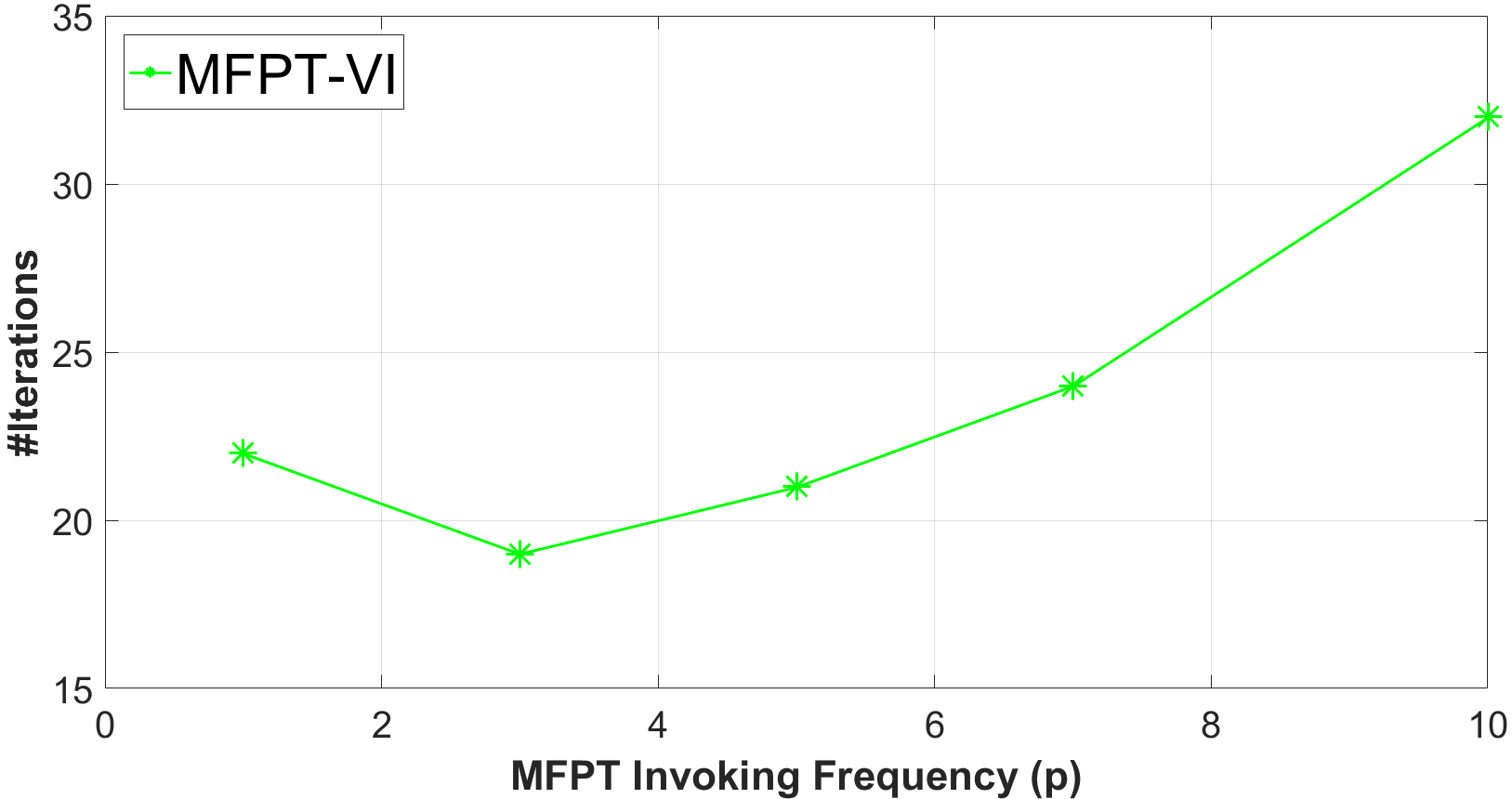}}  
	\caption{\small Trends of time and iterations under differing MFPT invoking frequencies. (a) The overall runtime of invoking MFPT with every $p$ VI iterations. (b) The number of iterations of invoking MFPT with every $p$ VI iterations.
	}
	\label{fig:TimeIter1}
	\vspace{-5pt}
\end{figure}

In Fig.~\ref{fig:TimeIter}, the number on the $x$-axis represents that the MFPT values are computed after every \textit{p} VI iterations. The time taken by MFPT-VI initially decreases to a certain point and then increases from there on as \textit{p} increases. 
We also observe a similar behavior in Fig. ~\ref{fig:ConvIter} when we analyze the number of iterations taken to converge under varying $p$.

This implies that, (1) invoking MFPT calculation at every iteration is not the optimal way; and (2) for different application scenarios with different patterns of transition matrices, the best range of $p$ can be obtained by running a few trials, using probing mechanisms such as the binary search. In our testing scenario, a $p$ ranging from three to five is a good choice.

\begin{figure}[ht]
  \centering
  \subfigure[]
        {\label{fig:3D_1}\includegraphics[height=1.3in]{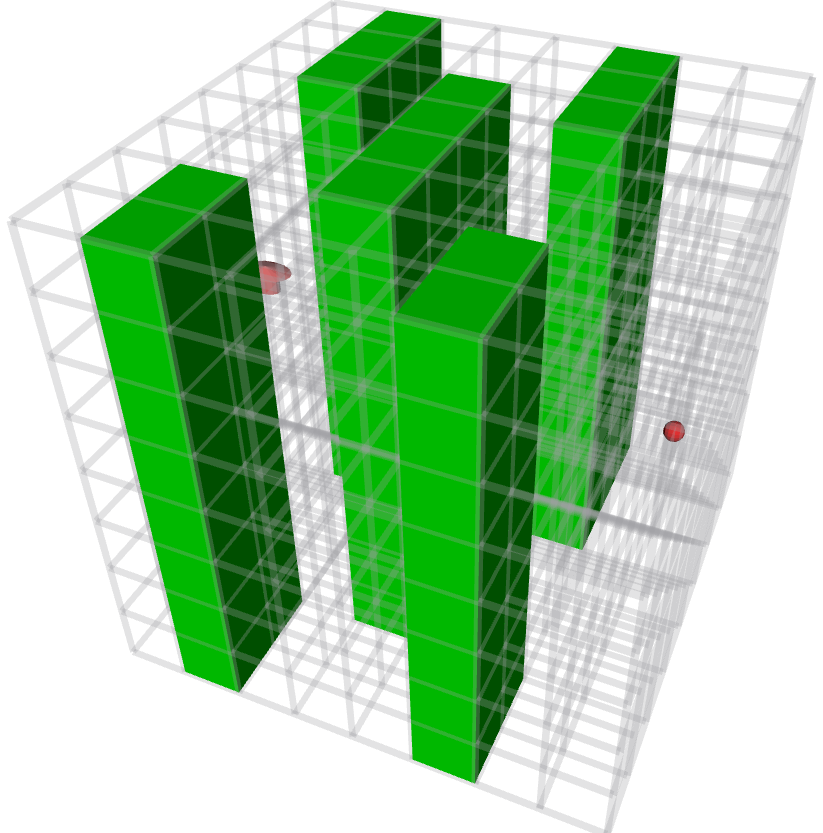}}
        \quad \quad 
  \subfigure[]
        {\label{fig:3D_3}\includegraphics[height=1.3in]{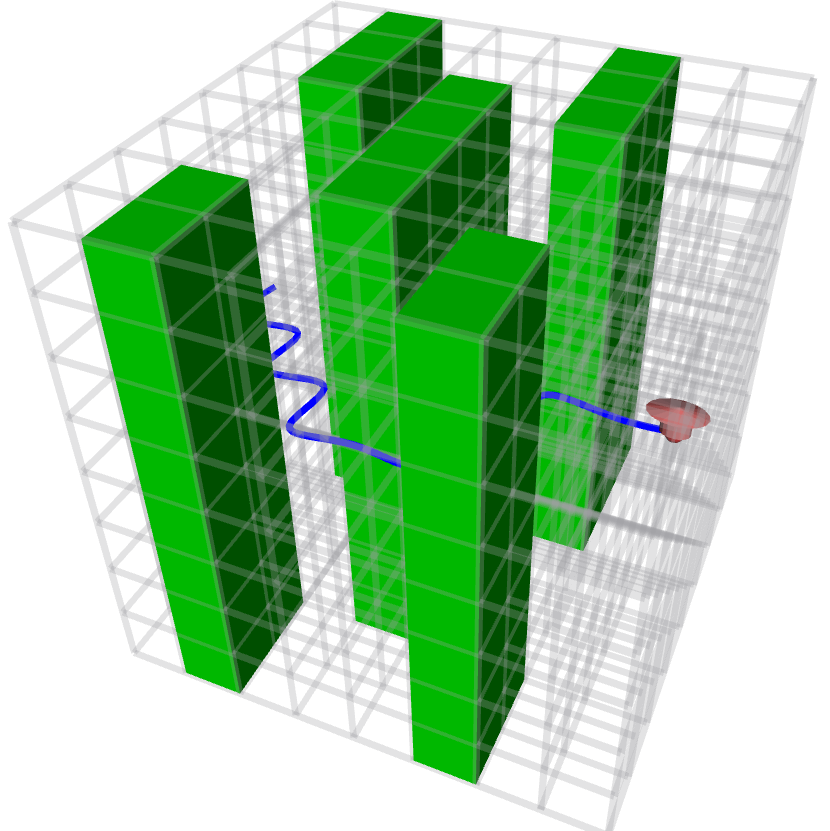}}
        \quad \quad 
	\vspace{-5pt}
	\caption{\small (a) Demonstration of simulation environment, with the agent's initial state (pink) and the goal state (red). Green blocks are obstacles; (b) A trajectory (blue) completed by the agent to reach the goal.}
	\label{fig:3DTrajectory}
	\vspace{-15pt}
\end{figure}

\subsection{3D Grid Setup}
A demonstration of the trajectory taken by the agent in 3D simulation environment is shown in Fig.~\ref{fig:3DTrajectory}. Next, we compare our proposed algorithms with their corresponding baseline algorithms in terms of the practical runtime performance and iterations required to converge.

\begin{figure}[t]
  \centering
  \subfigure[]
        {\label{fig:TS_V_3D}\includegraphics[height=1.3in]{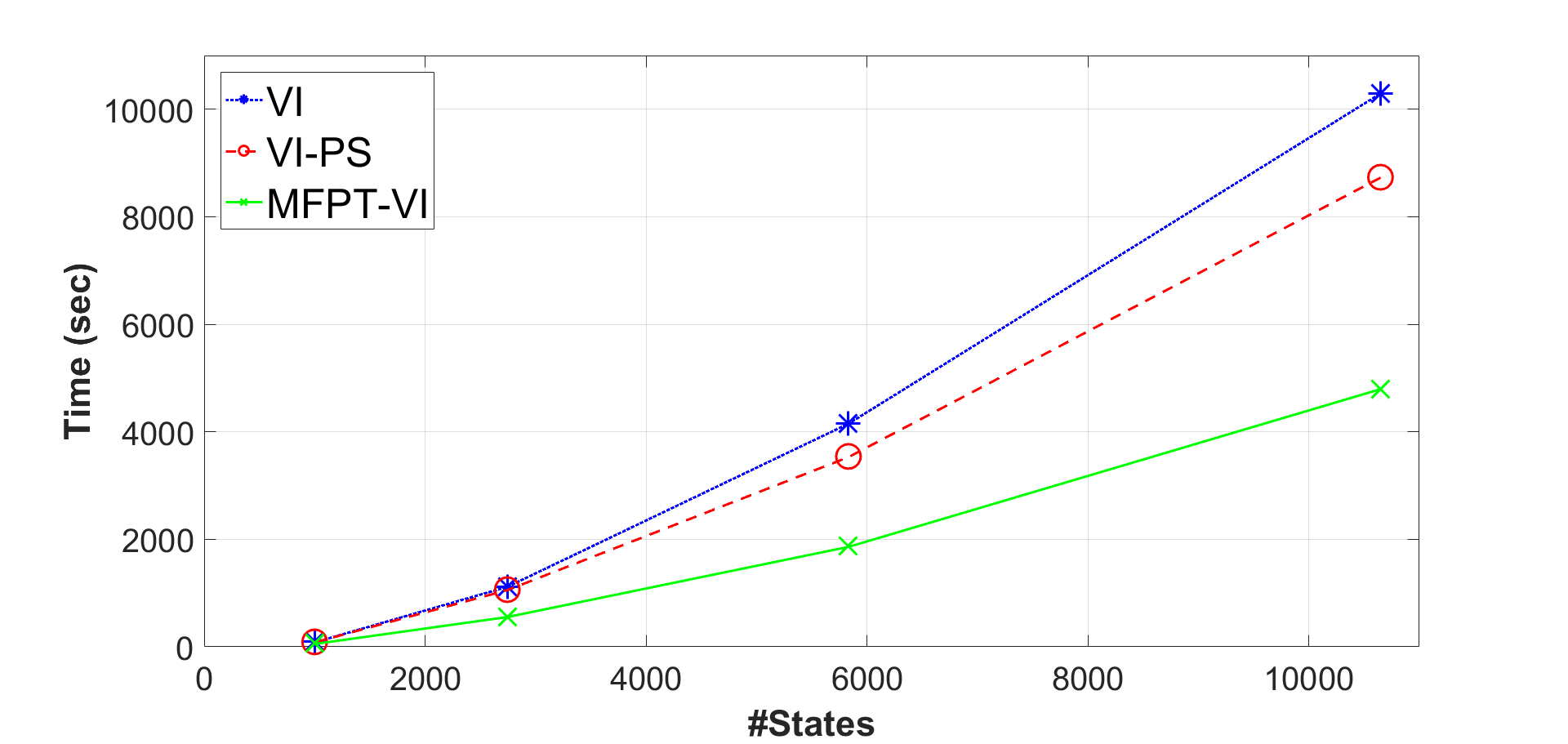}}
  \subfigure[]    
        {\label{fig:TS_P_3D}\includegraphics[height=1.3in]{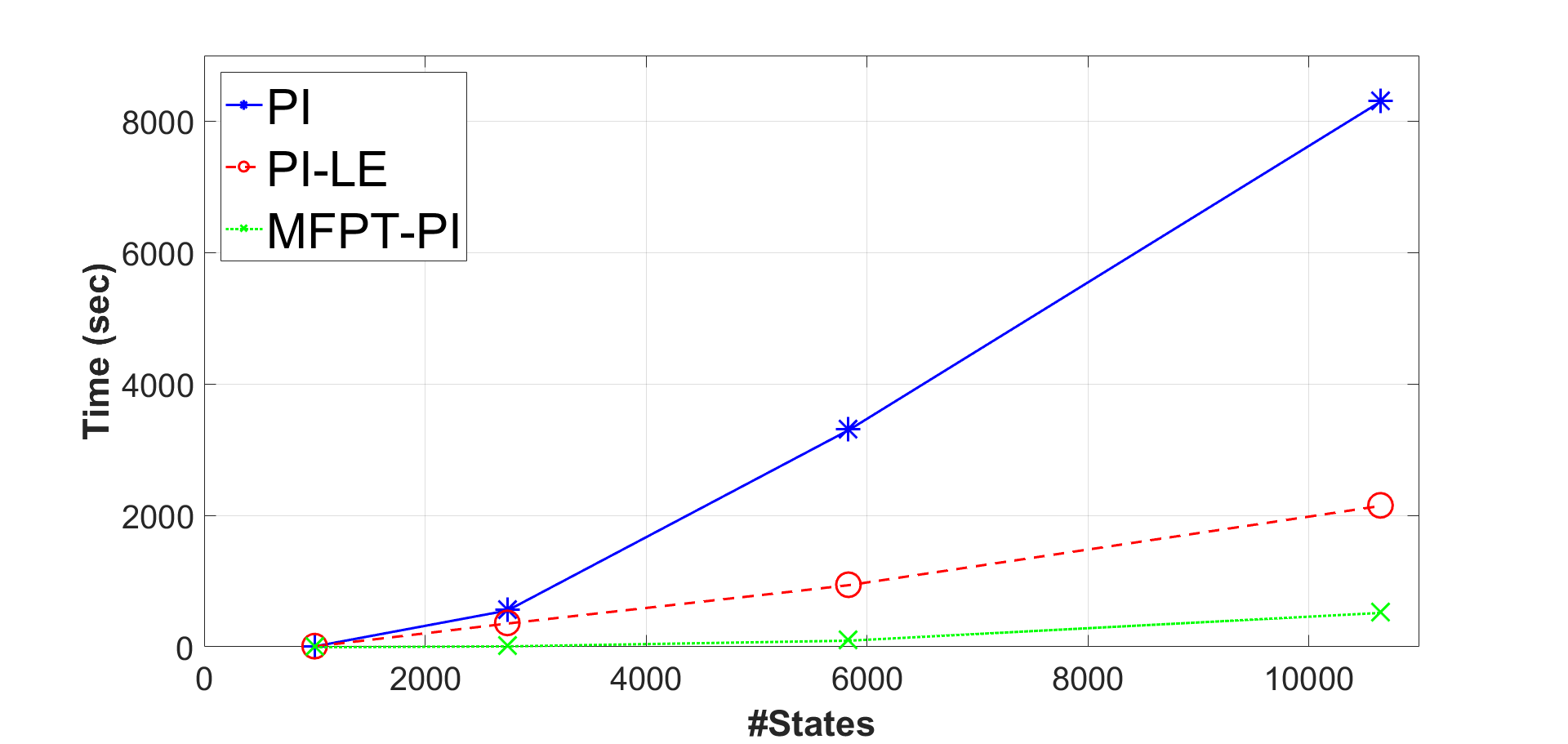}}  
	\vspace{-5pt}
	\caption{\small Time comparisons between the baseline methods and our proposed algorithms, with changing numbers of states ($x$-axis). (a) Variants of value iteration methods. (b) Variants of policy iteration methods.
	}
	\label{fig:TimeVsStates3D}
	\vspace{-10pt}
\end{figure}

We first investigate the time taken by the algorithms with changing number of states. 
Fig.~\ref{fig:TS_V_3D} 
compares the time taken by VI, VI-PS and MFPT-VI algorithms.
The results show that our proposed MFPT-VI is much faster than the VI and VI-PS. Also, as expected, due to the prioritized sweeping heuristic, VI-PS is faster than VI.
In Fig.~\ref{fig:TS_P_3D}, we compare the time taken by PI, PI-LE, and MFPT-PI. 
The results show that MFPT-PI is the fastest compared to the other two algorithms.

The outstanding performance of MFPT-VI and MFPT-PI indicate that, reachability characterization based framework is superior to state-of-the-art solutions, yet the implementation of our methods are as simple as the classic ones.

\begin{figure}[t]
  \centering
  \subfigure[]
        {\label{fig:CS_V_3D}\includegraphics[height=1.4in]{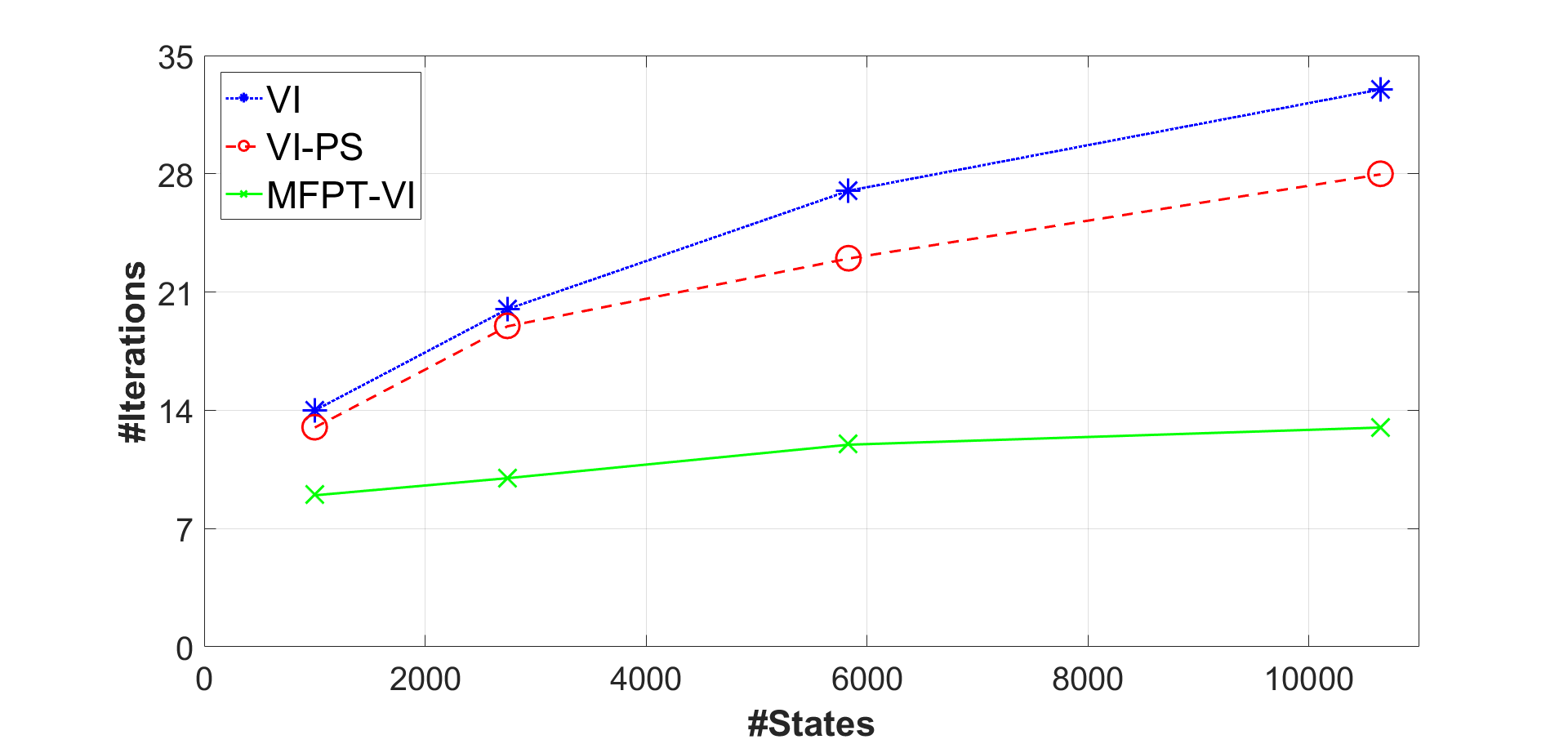}}
        \quad
  \subfigure[]    
        {\label{fig:CS_P_3D}\includegraphics[height=1.4in]{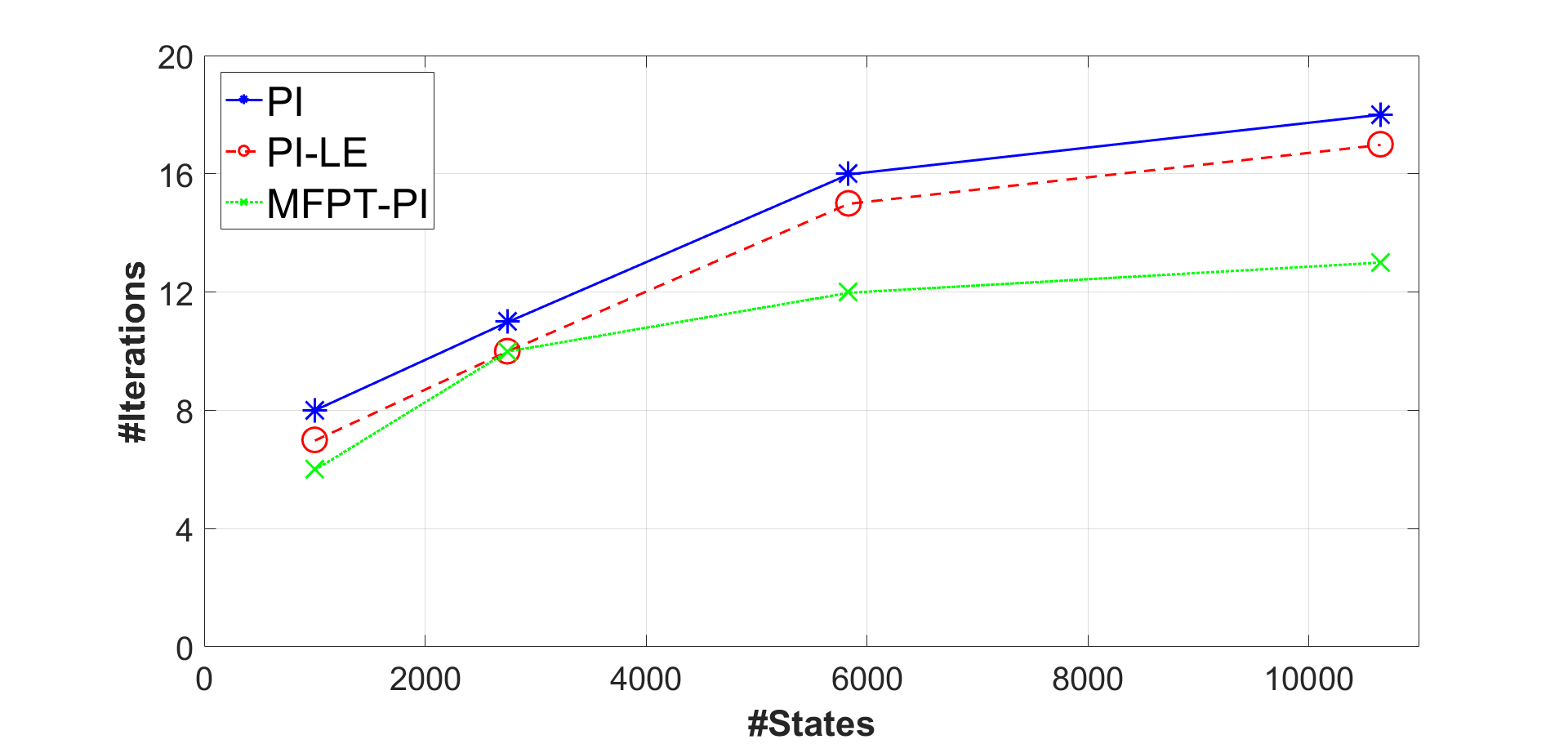}}	
    \vspace{-5pt}
	\caption{\small Iteration to convergence between the baseline methods and our proposed algorithms, with changing numbers of states ($x$-axis). (a) Variants of value iteration methods. (b) Variants of policy iteration methods.
	}
	\label{fig:ConvergenceVsStates3D}
	\vspace{-10pt}
\end{figure}

As in the case of 2D grid setup, we also analyze the number of iterations required by the algorithms to converge as the number of states change.
In Fig.~\ref{fig:CS_V_3D},we compare the number of iterations taken by VI, VI-PS and MFPT-VI, respectively. 
Similar to the 2D grid setup, our results reveal that MFPT-VI converges the fastest compared to VI and VI-PS. 
Fig.~\ref{fig:CS_P_3D} compares the iterations taken by PI, PI-LE, and MFPT-PI. 
Again, we can see that MFPT-PI converges the fastest among all three algorithms.

Another interesting observation is that as the number of states increase, the number of iterations required to converge flattens out for both MFPT-VI and MFPT-PI. These results clearly show the power of reachability characterization, which very well captures the convergence optimization feature, thereby requiring much fewer iterations to reach optimality.

\vspace{-4pt}
\section{Conclusions}
\label{conclusion}

In this paper, we propose a new framework for efficiently solving the MDPs.
The proposed method explores reachability of states using MFPT values, which characterizes the degree of difficulty of reaching given goal states. 
Different from the classic VI and PI methods, the proposed solutions based on MFPTs reflect a patterned landscape of states, which very well captures -- and also allows us to visualize -- the degree of importance of states.  
The reachability characterization enables one to design efficient heuristics such as value prioritization and policy approximation, and we propose two specific algorithms called MFPT-VI and MFPT-PI.
We show that the implementation of proposed methods is as simple as classic methods, but our algorithms converge much faster, with less runtime and fewer iterations than state-of-art approaches.


\vspace{-4pt}
{
\bibliographystyle{unsrt}
\bibliography{reference}

\begin{thebibliography}{10}

\bibitem{russell02}
Stuary Russell and Peter Norvig.
\newblock Artifical intelligence: A modern approach.
\newblock Accessed October 22, 2004 at http://aima.cs.berkeley.edu/, 2002.

\bibitem{sigaud2013markov}
Olivier Sigaud and Olivier Buffet.
\newblock {\em Markov decision processes in artificial intelligence}.
\newblock John Wiley \& Sons, 2013.

\bibitem{puterman2014markov}
Martin~L Puterman.
\newblock {\em Markov decision processes: discrete stochastic dynamic
  programming}.
\newblock John Wiley \& Sons, 2014.

\bibitem{white1993survey}
Douglas~J White.
\newblock A survey of applications of markov decision processes.
\newblock {\em Journal of the Operational Research Society}, 44(11):1073--1096,
  1993.

\bibitem{BoutilierDTP99}
Craig Boutilier, Thomas Dean, and Steve Hanks.
\newblock Decision-theoretic planning: Structural assumptions and computational
  leverage.
\newblock {\em Journal of Artificial Intelligence Research}, 11:1--94, 1999.

\bibitem{sutton1990integrated}
Richard~S Sutton.
\newblock Integrated architectures for learning, planning, and reacting based
  on approximating dynamic programming.
\newblock In {\em Proceedings of the seventh international conference on
  machine learning}, pages 216--224, 1990.

\bibitem{busoniu2010reinforcement}
Lucian Busoniu, Robert Babuska, Bart De~Schutter, and Damien Ernst.
\newblock {\em Reinforcement learning and dynamic programming using function
  approximators}, volume~39.
\newblock CRC press, 2010.

\bibitem{van2012reinforcement}
Martijn van Otterlo and Marco Wiering.
\newblock Reinforcement learning and markov decision processes.
\newblock In {\em Reinforcement Learning}, pages 3--42. Springer, 2012.

\bibitem{howarddynamic60}
Ronald~A. Howard.
\newblock {\em Dynamic Programming and {M}arkov Processes}.
\newblock MIT Press, Cambridge, MA, 1960.

\bibitem{Bertsekas1987}
D.~P. Bertsekas.
\newblock {\em Dynamic Programming: Deterministic and Stochastic Models}.
\newblock Prentice-Hall, Englewood Cliffs, N.J., 1987.

\bibitem{Moore93prioritizedsweeping}
Andrew~W. Moore and Christopher~G. Atkeson.
\newblock Prioritized sweeping: Reinforcement learning with less data and less
  time.
\newblock In {\em Machine Learning}, pages 103--130, 1993.

\bibitem{parr1998generalized}
David Andre, Nir Friedman, and Ronald Parr.
\newblock Generalized prioritized sweeping.
\newblock {\em Advances in Neural Information Processing Systems}, 1998.

\bibitem{wingate2005prioritization}
David Wingate and Kevin~D Seppi.
\newblock Prioritization methods for accelerating mdp solvers.
\newblock {\em Journal of Machine Learning Research}, 6(May):851--881, 2005.

\bibitem{boutilier2000stochastic}
Craig Boutilier, Richard Dearden, and Mois{\'e}s Goldszmidt.
\newblock Stochastic dynamic programming with factored representations.
\newblock {\em Artificial intelligence}, 121(1):49--107, 2000.

\bibitem{Barto1995}
Andrew~G. Barto, Steven~J. Bradtke, and Satinder~P. Singh.
\newblock Learning to act using real-time dynamic programming.
\newblock {\em Artif. Intell.}, 72(1-2):81--138, January 1995.

\bibitem{bonet2003labeled}
Blai Bonet and Hector Geffner.
\newblock Labeled rtdp: Improving the convergence of real-time dynamic
  programming.
\newblock In {\em ICAPS}, volume~3, pages 12--21, 2003.

\bibitem{andre2002state}
David Andre and Stuart~J Russell.
\newblock State abstraction for programmable reinforcement learning agents.
\newblock In {\em AAAI/IAAI}, pages 119--125, 2002.

\bibitem{li2006towards}
Lihong Li, Thomas~J Walsh, and Michael~L Littman.
\newblock Towards a unified theory of state abstraction for mdps.
\newblock In {\em ISAIM}, 2006.

\bibitem{kemeny1959finite}
John~G Kemeny, Hazleton Mirkill, J~Laurie Snell, and Gerald~L Thompson.
\newblock {\em Finite mathematical structures}.
\newblock Prentice-Hall, 1959.

\bibitem{AssafSharedShanthikumar1985}
David Assaf, Moshe Shared, and J.~George Shanthikumar.
\newblock First-passage times with pfr densities.
\newblock {\em Journal of Applied Probability}, 22(1):185--196, 1985.

\bibitem{Golub1996}
Gene~H. Golub and Charles~F. Van~Loan.
\newblock {\em Matrix Computations (3rd Ed.)}.
\newblock Johns Hopkins University Press, Baltimore, MD, USA, 1996.

\end{thebibliography}
}

\end{document}